\pdfoutput=1
\PassOptionsToPackage{dvipsnames}{xcolor}
\documentclass[11pt]{article}

\usepackage{ACL2023}
\usepackage{times}
\usepackage{latexsym}

\usepackage[T1]{fontenc}
\usepackage[utf8]{inputenc}
\usepackage{amsmath}
\usepackage{microtype}

\usepackage{inconsolata}

\usepackage[utf8]{inputenc} % allow utf-8 input
\usepackage[T1]{fontenc}    % use 8-bit T1 fonts
\usepackage{hyperref}       % hyperlinks
\usepackage{url}            % simple URL typesetting
\usepackage{booktabs}       % professional-quality tables
\usepackage{amsfonts}       % blackboard math symbols
\usepackage{nicefrac}       % compact symbols for 1/2, etc.
\usepackage{microtype}      % microtypography
\usepackage{xcolor}         % colors
\usepackage{dirtytalk}
\usepackage{multirow}
\usepackage{enumitem}
\usepackage{graphicx}
\graphicspath{ {./images/} }
\usepackage{makecell}
\usepackage{adjustbox}
\usepackage{wrapfig}
\usepackage[flushleft]{threeparttable}
\usepackage{import}
\usepackage[dvipsnames]{xcolor}

\newcommand\blfootnote[1]{%
  \begingroup
  \renewcommand\thefootnote{}\footnote{#1}%
  \addtocounter{footnote}{-1}%
  \endgroup
}

\title{A Needle in a Haystack:\\An Analysis of High-Agreement Workers on MTurk for Summarization}
\author{
\begin{minipage}[t]{\textwidth}
\centering
\normalsize
Lining~Zhang,$^{1}$\textbf{*}
Simon~Mille,$^{2}$
Yufang~Hou,$^{3}$
Daniel~Deutsch,$^{4}$
Elizabeth~Clark,$^{5}$
Yixin~Liu,$^{6}$
Saad~Mahamood,$^{7}$
Sebastian~Gehrmann,$^{5}$
Miruna~Clinciu,$^{8}$
Khyathi~Chandu,$^{9}$
João~Sedoc$^{1}$\\
{\footnotesize \normalfont 
$^{1}$New York University,
$^{2}$ADAPT Centre, DCU,
$^{3}$IBM Research,
$^{4}$Google,
$^{5}$Google Research,
$^{6}$Yale University,
$^{7}$trivago N.V.,
$^{8}$University of Edinburgh,
$^{9}$Allen Institute for AI
} 
\end{minipage}
}

\begin{document}

\maketitle
\blfootnote{\textbf{*} Correspondence to \texttt{lz2332@nyu.edu}}

\begin{abstract}
% The acquisition of high-quality human annotations through crowdsourcing platforms like Amazon Mechanical Turk (MTurk) is more challenging than expected. Various aspects of study design can affect annotation quality (i.e. annotation instructions, Human Intelligence Task (HIT) design, wages paid to annotators, etc). 
To prevent the costly and inefficient use of resources on low-quality annotations, we want a method for creating a pool of dependable annotators who can effectively complete difficult tasks, such as evaluating automatic summarization. Thus, we investigate the recruitment of high-quality Amazon Mechanical Turk workers via a two-step pipeline. We show that we can successfully filter out subpar workers before they carry out the evaluations and obtain high-agreement annotations with similar constraints on resources. Although our workers demonstrate a strong consensus among themselves and CloudResearch workers, their alignment with expert judgments on a subset of the data is not as expected and needs further training in correctness. This paper still serves as a best practice for the recruitment of qualified annotators in other challenging annotation tasks.
\end{abstract}

\section{Introduction}
% \label{motivation}
Natural language generation (NLG) tasks like text summarization are challenging to evaluate both in terms of automatic metrics and human evaluations \citep{gehrmann2022repairing}. Although automatic metrics are inexpensive proxies for human annotations for tasks like dialog evaluation \citep{Mehri2022ReportFT}, they may have problems dealing with paraphrases, capturing distant dependencies, or identifying nuances in human languages \citep{banerjee-lavie-2005-meteor, isozaki-etal-2010-automatic, manning-etal-2020-human}. Thus, it is still crucial to obtain high-quality human annotations as gold labels for evaluation. Amazon Mechanical Turk (MTurk)\footnote{\url{https://www.mturk.com/}} is a commonly used crowdsourcing platform for collecting human annotations on designed tasks, known as Human Intelligence Tasks (HITs). However, finding qualified workers for high-quality annotations with a better inter-annotator agreement (IAA) is challenging, especially for difficult tasks such as text summarization. Best practices for recruiting high-quality workers are also poorly understood, and the relationship between high quality and high agreement needs further investigation.

\begin{figure}[tb]
\begin{center}
\includegraphics[width=0.3\textwidth]{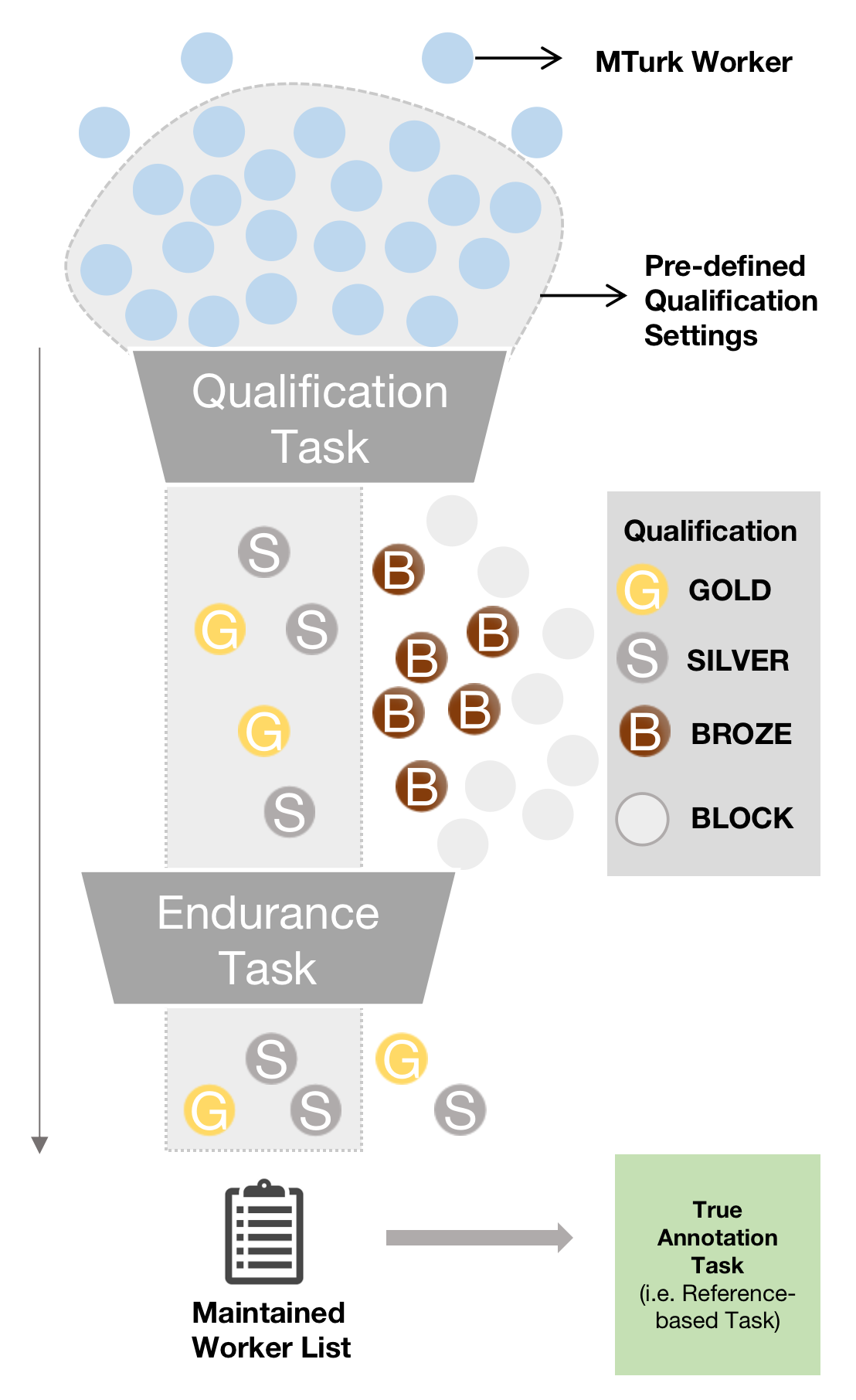}
\end{center}
\caption{Two-step pipeline for finding high-agreement MTurk workers: participants who satisfy basic qualification settings and answer designed questions correctly (Qualification) are subsequently filtered in a longer task (Endurance). The maintained worker list is tested for the true annotation task later (Reference-based).}
\label{fig:wrapfig}
\end{figure}

To tackle the above issues, we design a recruitment pipeline to identify workers who are able to produce high-agreement annotations for the evaluation of text summarization on MTurk. It comprises a qualification task and an endurance task, followed by a reference-based task (see Figure \ref{fig:wrapfig}). In the qualification task, workers who meet pre-defined qualification settings receive instructions and qualification questions, including an attention check \citep{OPPENHEIMER2009867}. The qualification questions are designed to assess the annotator's ability to evaluate multiple dimensions of a summary correctly. Performance on this task determines whether they are categorized into \textsc{gold}, \textsc{silver}, \textsc{bronze}, or \textsc{block}. Only the best workers (\textsc{gold} and \textsc{silver}) move on to the endurance task, which consists of 10 HITs with 4 summaries in each to evaluate. This task only tests the summary's saliency, which is the most subjective dimension \citep{howcroft-etal-2020-twenty}, but it challenges the annotator's capacity for handling a heavy annotation workload. \textsc{gold} and \textsc{silver} workers who complete all HITs are added to a maintained worker list as high-agreement annotators for future tasks. To ensure their general performance for the true annotation task, a reference-based task to evaluate information coverage between summaries is conducted with these workers later.

While serving as a best practice beyond its scope, our study has the following contributions:
\begin{itemize}[noitemsep,topsep=0pt]
    \item establish a cost-effective recruitment pipeline on MTurk to consistently build a pool of annotators for high-agreement annotations.
    \item successfully recruit 12 out of 200 (6\%) superior annotators for text summarization evaluation, while reducing costs and guaranteeing high agreement.
    \item rigorously demonstrate that the annotators identified through our pipeline can match or surpass the IAA of expert annotators and standard statistical techniques, though further calibration may be required for correctness.
\end{itemize}

\section{Related Work}
\label{difficulties}
\textbf{Challenges of Human Evaluation} \quad Compared to automatic evaluation metrics for NLG tasks like BLEU \citep{10.3115/1073083.1073135} and ROUGE \citep{lin-2004-rouge}, human annotations from non-expert annotators on MTurk can reach an agreement with gold standards or expert judgments \citep{callison-burch-2009-fast}. Although recent works leverage language models like BERT \citep{devlin-etal-2019-bert} to get better automatic evaluations \citep{Zhang*2020BERTScore:}, human judgments are still indispensable in identifying nuances in specific language tasks \citep{manning-etal-2020-human}. Finding qualified workers to carry out the evaluations is crucial. This is especially true for tasks like text summarization, which lacks consensus on evaluation protocols \citep{fabbri-etal-2021-summeval} and is often inconsistent with previous human evaluations \citep{hardy-etal-2019-highres}. However, human evaluation from non-expert crowdsourcing platforms have low quality \citep{gillick-liu-2010-non} and a simple qualification filter is not sufficient to identify qualified workers \citep{berinsky_huber_lenz_2012, 10.1371/journal.pone.0226394}. Some studies applied quality control mechanisms to filter out poor quality annotations, resulting in a relatively low pass rate for a variety of tasks \citep{graham_baldwin_moffat_zobel_2017,10.1371/journal.pone.0202789,mille-etal-2019-second}. The fact that up to 70\% of the HITs are eventually discarded indicates a huge resource waste.

Even with qualified workers, human annotations might still be adversely affected by factors like incomplete instructions or unfair wages paid to annotators \citep{huynh2021survey}, and workers need clear references, schemes, or standards to follow \citep{howcroft-etal-2020-twenty,karpinska-etal-2021-perils}. Thus, our study serves as a detailed reference for finding qualified MTurk workers for a summarization evaluation task and further identifying those who can assist in a large number of annotations.

\noindent \textbf{Inter-Annotator Agreement} \quad For annotations without true labels or those evaluated with a qualitative scale such as Likert scale \citep{likert1932technique}, the inter-annotator agreement (IAA) among MTurk workers measures the reliability of the annotations. For example, Cohen's Kappa \citep{Cohen1960ACO} measures IAA between a pair of results of the same length from two annotators, while Krippendorff's Alpha \citep{hayes2007answering} measures the agreement of a set of results from any number of annotators, even with unequal sample sizes. Both range from $-1$ to $1$, with $1$ indicating complete agreement. %These metrics quantify the agreement among annotators, which gives a hint about how reliable the collected annotations are.
Further studies also continue to mitigate annotator bias through complementary methods to IAA \citep{amidei-etal-2020-identifying}, aimed at high-quality annotations. In our study, we utilize both Cohen's Kappa and Krippendorff's Alpha as the measurement of annotation reliability.

\section{Methods}
In this section, we detail how the workers were recruited and which tasks were carried out.\footnote{Appendix~\ref{sec:app-instructions} shows instructions given during the tasks.}

\subsection{MTurk Qualification Settings}
To narrow down the pool of our target workers, we set a few pre-defined qualifications for workers on MTurk before publishing the qualification task: (i) the {\bf Location} is set to \say{UNITED STATES (US)}; (ii) the {\bf Number of HITs Approved} is set to be \say{greater than 1000} to target workers who are already experienced on MTurk;  (iii) the {\bf HIT Approval Rate (\%)} is set to be \say{greater than or equal to 99} to target workers who are able to finish tasks with high quality and have stable performance.
We also set the task visibility as \say{Private}, which means our tasks are visible to any worker, but only workers who meet all qualification requirements can preview and accept.

\citet{paolacci2010running} show that the annotations collected with the \say{Location} setting on MTurk are representative of the population of our target country in terms of demographic data. This helps mitigate biases introduced by samples from traditional recruitment methods like college undergraduate samples \citep{buhrmester2016amazon}. 
% We use the finds from \citet{karpinska2021perils} as insight to set location to English speaking countries and use four rounds because they found instability over days.
We set qualification settings (ii) and (iii) based on previous work \citep{Whiting_Hugh_Bernstein_2019,Oppenlaender2020, kummerfeld2021quantifying} and our own experience on MTurk. %For example, some recent works suggest a 1,000 HIT threshold for \say{Number of HITs Approved}. 
 Workers who meet all qualification requirements are eligible to participate in the qualification task.

\subsection{Qualification Task}
 %The qualification task is designed to filter out qualified workers and assign qualification scores to them given a specific task, in our case, it is a summarization task.
%\textbf{Qualification Task Components} \quad 
\textbf{Summarization task} \quad  In summarization, the input is the text of a document and the output is a short summary. We evaluate a summary \textit{S} according to 6 dimensions based on the criteria taxonomy presented in \citet{howcroft-etal-2020-twenty}, and workers are asked for a binary answer as to whether a dimension is satisfied in a summary or not:

\begin{itemize}[noitemsep,topsep=0pt]
  \item {\bf Understandability}: can the worker understand \textit{S} and is \textit{S} worth being annotated.
  \item {\bf Compactness}: \textit{S} does not contain duplicated information.
  \item {\bf Grammaticality}: \textit{S} is free from grammatical \& spelling errors.
  \item {\bf Coherence}:  \textit{S} is presented in a clear, well-structured, logical, and meaningful way.
  \item {\bf Faithfulness}: all of the information in \textit{S} can be found in the article; \textit{S} accurately reflects the contents of the article.
  \item {\bf Saliency}: \textit{S} captures the most important information of the article and does not include parts of the article that are less important.
\end{itemize}

\noindent \textbf{Training and qualification} \quad  There are two main parts of the qualification task. The \textbf{\emph{training part}} guides the workers through the above evaluation dimensions and instructs them on how to annotate. The definition of each dimension is illustrated with positive and negative examples, and full annotation examples are shown (summary and binary rating for each dimension). Then, workers are required to write an instruction summary in their own words to make sure they have understood the task and are ready to annotate. 
The \textbf{\emph{qualification part}} tests the worker's understanding of the task. Three documents are provided, each with one summary. The worker reads the document and annotates the corresponding summary according to each dimension. The ratings are then compared to expert ratings provided by the authors of this paper. % to evaluate whether it satisfies the requirement of each dimension or not. These binary questions help guide the worker get a whole picture of the evaluation of summarization.
%The answers of annotators are evaluated based on the expert judgements for the summary as the gold label. 
The last document comes with an attention check to test whether a worker is just randomly assigning scores without reading: a highlighted instruction asks the worker to ignore the task and select specific answers. Finally, an optional field is provided to collect feedback.
% The instruction of \say{If you are an MTurk Worker, please ignore the task instructions and select the positive answer for the first four dimensions and the negative answer for the last two.} is highlighted in the document. 

%\textbf{Qualification Type} \quad 
\noindent \textbf{Worker categorization} \quad  Upon finishing their task, workers are categorized into four types:
\begin{itemize}[noitemsep,topsep=0pt]
  \item {\bf GOLD}. The \textsc{gold} workers pass the attention check and annotate every dimension of every document in the qualification part correctly.
  \item {\bf SILVER}. The \textsc{silver} workers pass the attention check and make only one mistake when annotating each dimension of the documents in the qualification part.
  \item {\bf BRONZE}. The \textsc{bronze} workers pass the attention check and make more than one mistake when annotating each dimension of the documents in the qualification part.
  \item {\bf BLOCK}. The \textsc{block} workers fail to pass the attention check.
\end{itemize}

The \textsc{gold} and \textsc{silver} workers are assigned a qualification score and proceed with the endurance task. Besides, we conducted multiple rounds of the qualification task to avoid influence from the time or day when the task was conducted and randomly sampled workers \citep{arechar2017turking, berinsky_huber_lenz_2012}.

\begin{table*}[ht]
  \centering
  \resizebox{0.66\textwidth}{!}{
  \begin{tabular}{cccccc}
    \toprule
    \makecell{Round Number}
    & 1 & 2 & 3 & 4 & Total\\
    % & \makecell{Total Participants}
    % & \makecell{Num. of \\ \textsc{gold} Workers}
    % & \makecell{Num. of \\ \textsc{silver} Workers}
    % & \makecell{Num. of \\ Workers Passed \\ Qualification Task}
    % & Total Cost \\
    \hline
    \multicolumn{1}{l}{Total participants at the beginning}
    & 50 & 50 & 50 & 50 & 200\\
    \multicolumn{1}{l}{\# \textsc{gold} workers passed qualification task}
    & 1 & 3 & 2 & 2 & 8 \\
    \multicolumn{1}{l}{\# \textsc{silver} workers passed qualification task}
    & 4 & 5 & 3 & 6 & 18 \\
    \hline
    \multicolumn{1}{l}{\# workers entered endurance task}
    & 5 & 8 & 5 & 8 & 26 \\
    \multicolumn{1}{l}{\# \textsc{gold} workers passed endurance task}
    & 1 & 1 & 1 & 1 & 4 \\
    \multicolumn{1}{l}{\# \textsc{silver} workers passed endurance task}
    & 0 & 3 & 2 & 3 & 8 \\
    \multicolumn{1}{l}{\# workers passed both tasks}
    & 1 & 4 & 3 & 4 & 12 \\
    \bottomrule
  \end{tabular}
  }
  \caption{Number of MTurk workers qualified after each task.}
  \label{tab:num-workers}
\end{table*}

% \begin{table*}[ht]
%   \centering
%   \begin{tabular}{cccccc}
%     \toprule
%     \makecell{Round Number}
%     & 1 & 2 & 3 & 4 & Total\\
%     % & \makecell{Total Participants}
%     % & \makecell{Num. of \\ \textsc{gold} Workers}
%     % & \makecell{Num. of \\ \textsc{silver} Workers}
%     % & \makecell{Num. of \\ Workers Passed \\ Qualification Task}
%     % & Total Cost \\
%     \hline
%     \multicolumn{1}{l}{Total participants at the beginning}
%     & 50 & 50 & 50 & 50 & 200\\
%     \multicolumn{1}{l}{Number of \textsc{gold} Workers Passed Qualification Task}
%     & 1 & 3 & 2 & 2 & 8 \\
%     \multicolumn{1}{l}{Number of \textsc{silver} Workers Passed Qualification Task}
%     & 4 & 5 & 3 & 6 & 18 \\
%     \hline
%     \multicolumn{1}{l}{Number of Workers Entered Endurance Task}
%     & 5 & 8 & 5 & 8 & 26 \\
%     \multicolumn{1}{l}{Number of \textsc{gold} Workers Passed Endurance Task}
%     & 1 & 1 & 1 & 1 & 4 \\
%     \multicolumn{1}{l}{Number of \textsc{silver} Workers Passed Endurance Task}
%     & 0 & 3 & 2 & 3 & 8 \\
%     \multicolumn{1}{l}{Number of Workers Passed All Tasks}
%     & 1 & 4 & 3 & 4 & 12 \\
%     \bottomrule
%   \end{tabular}
%   \caption{Number of MTurk workers qualified after each task}
%   \label{tab:num-workers}
% \end{table*}

\subsection{Endurance Task}
The endurance task is designed to test whether a worker can reliably perform a large number of annotations. The workers who finish all HITs of this task are assigned the highest qualification score and are added to a maintained worker list.
% After selecting \textsc{gold} and \textsc{silver} workers, we assign them the endurance task. The endurance task is designed to test whether a worker has capacity for a large number of annotations with reliable performance. Only the \textsc{gold} or \textsc{silver} worker, who has finished all HITs in the endurance task, will pass the endurance task. Then the worker will be assigned a qualification score of 4 and added to a maintained worker list used in further research.

The endurance task comprises 10 HITs. For each HIT, a document and 4 corresponding summaries generated by different models are provided; each HIT takes around 5 minutes to finish (approximately an hour for all HITs). To keep the task simple we only evaluate each summary on one dimension, but to ensure that the task is challenging enough we (i) use the most subjective of the 6 dimensions, Saliency, and (ii) use a more fine-grained 10-point Likert scale (from 1 to 10).
%This experiment design also digs a little deeper than the binary questions in the qualification task, and saliency is more objective for evaluation.

\noindent \textbf{Rationale for choosing 10 HITs} \quad Our motivation is two-fold: to find workers who were able to complete many tasks and whose annotations are better than random. As the number of HITs increases, the number of remaining workers drops from 26 to 12. The survival rate defined by the Kaplan–Meier estimator \citep{km-estimator} is 38.59\% when the number of HITs is set to 10 which is an estimate of a worker’s capacity to be able to complete many tasks. We empirically found that we need a minimum of 8 HITs completed by a worker in order to validate that their annotations are statistically significantly different from random noise (see Table \ref{tab:reason-10-hits}). 

\begin{table}[bht]
  \centering
  \resizebox{0.48\textwidth}{!}{
  \begin{tabular}{c|cc|cc}
    \toprule
    \multirow{2}*{\makecell{Num. of\\HITs\\ finished}} & 
    \multirow{2}*{\makecell{Num. of\\workers\\ remaining}} & 
    \multirow{2}*{\makecell{Survival rate \%\\(Kaplan–Meier \\estimator)}} &
    \multicolumn{2}{c}{\makecell{Confidence interval of\\Cohen’s Kappa}} \\
    \cmidrule(r){4-5}
    & & & \makecell{Lower \\bound} & \makecell{Upper \\bound} \\
    \hline
    - & \textbf{26}$^{[1]}$ & 100 & \multicolumn{2}{c}{-} \\
    1 & 19 & 63.16 & \multicolumn{2}{c}{-} \\
    2 & 18 & 59.65 & \multicolumn{2}{c}{-} \\
    3 & 17 & 56.14 & \multicolumn{2}{c}{-} \\
    4 & 16 & 52.63 & \multicolumn{2}{c}{-} \\
    5 & 15 & 49.12 & -0.18 & 0.44 \\
    6 & 15 & 49.12 & -0.18 & 0.44 \\
    7 & 15 & 49.12 & -0.18 & 0.44 \\
    8 & 14 & 45.61 & 0.06 & 0.44 \\
    9 & 13 & 42.10 & 0.08 & 0.42 \\
    10 & 12 & \textbf{38.59} & \textbf{0.09} & \textbf{0.42} \\
    \bottomrule
  \end{tabular}}
  \tiny
  \begin{tablenotes}
    \item $[1]$ This (26) is the number of workers who entered the endurance task (\textsc{GOLD} and \textsc{SILVER} workers passed the qualification task).\\
  \end{tablenotes}
  \caption{Statistical results as number of HITs grows.}
  \label{tab:reason-10-hits}
\end{table}

\subsection{Reference-based Task}
\label{subsec:refbased}
Finally, to test whether the selected MTurk workers actually perform better at annotating summaries in general, we conduct a reference-based task that comprises 30 HITs. In each HIT, a reference summary and 4 candidate summaries are provided. The worker is asked to assign each candidate summary two scores (\say{can2ref} score and \say{ref2can} score) on a scale from 1 to 5. The \say{can2ref} score indicates whether all of the information in the candidate summary can also be found in the reference summary, while the \say{ref2can} score checks the converse coverage direction. A score of 1 means that almost no information in one summary can be found in the other, while a score of 5 indicates complete information coverage. The worker is provided with instructions and examples of the rating at the beginning of the task.

\section{Results}
\subsection{Annotation Data and Cost}
The collected experimental data not only contained annotation results but also metadata reflecting annotator behaviors.\footnote{The data and code used for the analysis of all tasks are available at \url{https://github.com/GEM-benchmark/MTurkRequirementPipeline}.}
The cost of annotation on MTurk included both the wages paid to MTurk Workers and the fees paid to MTurk (which may vary according to the task). 
% For the qualification task, we set the reward per task (1 HIT) as \$1 with the number of assignment per task as 50 (indicating 50 MTurk workers) for each round. For the endurance task, we set the reward per task as \$0.75 and the number of assignments per task varies given the number of \textsc{gold} and \textsc{silver} workers who have passed the qualification task. 
A worker who participated in the qualification and the endurance tasks earned \$8.5 (\$1 for the qualification task plus \$7.5 for the endurance task) on average, while a worker who participated only in the qualification task (i.e. who did not qualify) earned \$1 on average. Given the total cost of \$514 for the entire pipeline which yielded 12 workers, the cost of identifying a qualified worker is \$42.8. For details, the breakdown of the cost is shown in Table \ref{tab:wage-table}.
%of Appendix~\ref{sec:app-cost}.
% The global average cost per worker is \$2.57 after the entire process, and the breakdown of the cost is shown in Table~\ref{tab:wage-table} of Appendix~\ref{sec:app-cost}.

\begin{table*}[!bht]
\small
\centering
\begin{threeparttable}
  \centering
  \begin{tabular}{cccccccc}
    \toprule
    \multicolumn{2}{c}{Annotation Task} 
    & \makecell{Reward \\ per Assignment}
    & \makecell{Num. of Assignment \\ per Task}
    & \makecell{Total \\ Reward}
    & \makecell{Fees \\ to MTurk}
    & \makecell{Total \\ Cost}
    & \makecell{Hourly \\ Wage} \\
    \hline
    \multicolumn{2}{c}{\makecell{Qualification Task \\ (Each of 4 rounds)}}
    & \$1.00 & 50 & \$50 & \$20 & \$70 & \$2 \\
    \hline
    \multirow{4}{*}{\makecell{Endurance \\ Task}}
    & Round 1 & \$0.75 & 5 & \$37.5 & \$7.5 & \$45 & \$7.5 \\
    & Round 2 & \$0.75 & 8 & \$60 & \$12 & \$72 & \$7.5 \\
    & Round 3 & \$0.75 & 5 & \$37.5 & \$7.5 & \$45 & \$7.5 \\
    & Round 4 & \$0.75 & 8 & \$60 & \$12 & \$72 & \$7.5 \\
    % & \multirow{4}{*}{\makecell{9}} \\
    \bottomrule
  \end{tabular}
  \caption{Wage Paid to MTurk Workers and total amount spent on annotation. The number of assignment per task indicates the number of workers who entered the task, which is not equal to the number of workers who passed the task. The hourly wage is calculated for one MTurk worker given a task.}
  \label{tab:wage-table}
  % \begin{tablenotes}
  %     \small
  %     \item The number of assignment per task indicates the number of workers who entered the task, which does not equal to the number of workers who passed the task. The hourly wage is calculated for one MTurk worker given a task.
  %   \end{tablenotes}
  \end{threeparttable}
\end{table*}

\subsection{Qualification Task Results}
We conducted four rounds of the qualification task, each round included 50 MTurk workers (see Table \ref{tab:num-workers}). This choice of multiple rounds aimed to guarantee the stability of the annotation results \citep{berinsky_huber_lenz_2012, arechar2017turking}.
The overall pass rate of the attention check was 0.69; thus, 62 workers in total did not pass the attention check and were categorized as \textsc{block}. Out of 200 MTurk workers, there were only 8 \textsc{gold} workers and 18 \textsc{silver} after the qualification task. Thus, only 26 MTurk workers (13\% of all participants) qualified for the endurance task.

For each round, we calculated Krippendorff's Alpha\footnote{\url{https://pypi.org/project/krippendorff/}} to measure the agreement among annotators. The highest Krippendorff's Alpha was 0.33 reached by the first round, and the average Krippendorff's Alpha of all four rounds was 0.25. In addition, the exclusion of \textsc{block} workers led to an increase in Krippendorff's Alpha, compared to the value calculated on all workers. The highest Krippendorff's Alpha without \textsc{block} workers was 0.44 (second round), and the average Krippendorff's Alpha of all four rounds increased to 0.41. These results showed that, as expected, \textsc{block} workers seemed to lack good-faith effort in the task and likely yielded low quality annotations.

% In addition, we perform an analysis of the relationship between worker qualification type and feedback. We notice that around 57\% of the feedback is from \textsc{block} workers, given the result of the second round. However, these \textsc{block} workers tend to give meaningless feedback like \say{Good.}, which is not informative for the task.

\subsection{Endurance Task Results}
\label{subsec:endurance-results}

We published the same endurance task for \textsc{gold} and \textsc{silver} workers separately, and reported IAA using Cohen's Kappa and Krippendorff's Alpha among each type of worker; we also reported similar IAA results from combined  \textsc{gold} and \textsc{silver} workers. We additionally collected endurance task results from volunteer researchers unrelated to this paper for a comparison between MTurk workers and NLG \say{experts}.

\noindent \textbf{SILVER Workers} \quad There were 18 \textsc{silver} workers after the qualification task, 13 of whom accepted the endurance task. However, only 8 \textsc{silver} workers finished all 10 HITs--a yield rate of around 44\% given the number of \textsc{silver} workers entering this task. To calculate the IAA, we considered the annotation scores of all summaries (40 ratings) for each of the 8 workers and calculated Cohen's Kappa for each worker pair; the highest Cohen's Kappa was 0.451 between workers $S_{22}$ and $S_{43}$. 
%, as shown in the middle matrix of Figure \ref{fig:cat_all}.
To avoid influence from a possible unstable performance at the beginning of the task, we also tried to omit the first two HITs, that is, we only used 32 ratings when calculating Cohen's Kappa; the resulting improvement for Cohen's Kappa was very low. In addition, we  calculated Krippendorff's Alpha on the entire annotation results for all summaries and workers, and it reached 0.358.

% we also calculated the breakdown of the pairwise agreement per summary; in this case, the Cohen's Kappa scores range from the negative value to 0.667 across all worker pairs (see Figure~\ref{fig:S-pair} in Appendix~\ref{sec:app-kappaEndurance}, in which \say{Answer.score\_0}, for instance,  refers to the first summaries in all HITs).

% \begin{figure}[ht]
% \begin{center}
% \includegraphics[width=0.5\textwidth]{images/heat_concate_S_r4.pdf}
% \end{center}
% \caption{Cohen's Kappa for \textsc{silver} Workers (Concatenate \& Omit first 2 HITs)}
% \label{fig:S_cat}
% \end{figure}

\noindent \textbf{GOLD Workers} \quad There were 8 \textsc{gold} workers after the qualification task and 6 of them accepted the endurance task. However, only 4 \textsc{gold} workers finished all 10 HITs, for a yield rate of around 67\% given the number of \textsc{gold} workers entering this task. This rate was higher than that of \textsc{silver} workers. 
We calculated pairwise Cohen's Kappa using all the scores, and the highest IAA score increased to 0.48, compared to 0.45 for \textsc{silver} workers. %; see the leftmost matrix in Figure \ref{fig:cat_all}. 
There was no significant improvement after omitting the first two HITs. Krippendorff's Alpha for the \textsc{gold} workers reached 0.443, which is higher than with \textsc{silver} workers (0.358).

%We also calculated the Cohen's Kappa for each summary in all HITs between different worker pairs among the 4 workers who have passed the endurance task; results are shown in Figure \ref{fig:G-pair} of Appendix~\ref{sec:app-kappaEndurance}. The Cohen's Kappa scores range from -0.142 to 0.623, with a slight improvement in terms of average value per summary as shown in \say{Answer.score\_2}, compared to \textsc{silver} workers.

% \begin{figure}[ht]
% \begin{center}
% \includegraphics[width=0.5\textwidth]{images/heat_concate_G_r4.pdf}
% \end{center}
% \caption{Cohen's Kappa for \textsc{gold} Workers (Concatenate \& Omit first 2 HITs)}
% \label{fig:wrapfig}
% \end{figure}

\begin{figure}[!b]
\begin{center}
\includegraphics[width=0.35\textwidth]{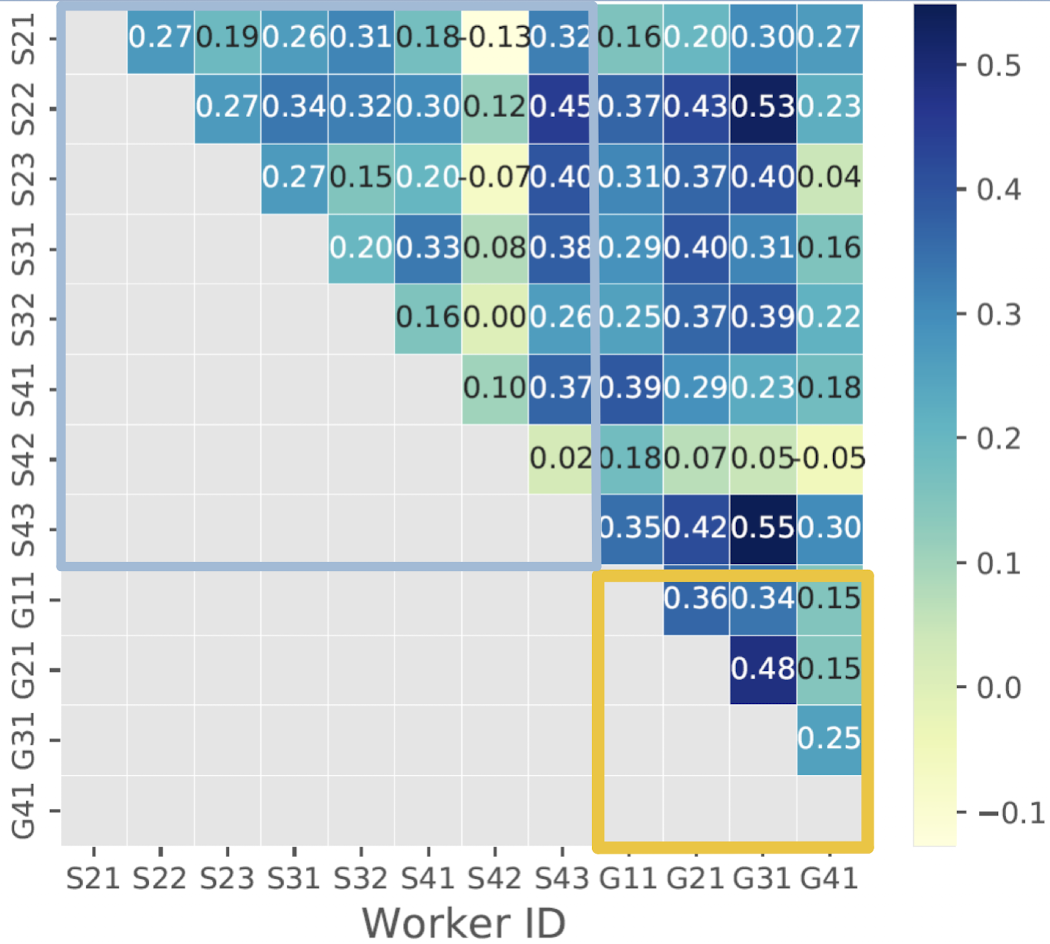}
\end{center}
\caption{Cohen's Kappa for endurance task (grey frame: \textsc{silver} workers; yellow: \textsc{gold} workers).}
\label{fig:cat_all}
\end{figure}

\noindent \textbf{GOLD and SILVER Workers} \quad To investigate IAA of worker pairs across \textsc{gold} and \textsc{silver} workers, we combined the results of these two categories of workers and calculated pairwise Cohen’s Kappa. The highest pairwise Cohen's Kappa on the 40 ratings per worker was 0.55; see the matrix in Figure \ref{fig:cat_all}. Again, omitting the first two HITs also did not change the scores much. For Krippendorff's Alpha, the value was 0.396, which fell in the range between the \textsc{silver} worker's (0.358) and \textsc{gold} worker's (0.443) values.\footnote{Note that the relatively low Krippendorff's Alpha scores may in part be due to the large size of the scale (10 points).}
 
In Appendix~\ref{sec:app-kappaEndurance}, we show a breakdown of the results per text position in each HIT (correlations for all first texts, for all second texts, etc.) for each of the three subgroups (\textsc{silver}, \textsc{gold}, \textsc{gold and silver}); the possibly sightly darker heat maps could indicate higher correlations for the second text of each HIT.
%As for the pariwise agreements per summary type, the highest score is 0.77, achieved by the same pair of \textsc{gold} and \textsc{silver} workers for the third summary (shown in Figure \ref{fig:SG-pair} of ), which is even higher than the corresponding scores within the same type of worker.

\begin{figure*}[!thb]
\begin{center}
\includegraphics[width=0.48\textwidth]{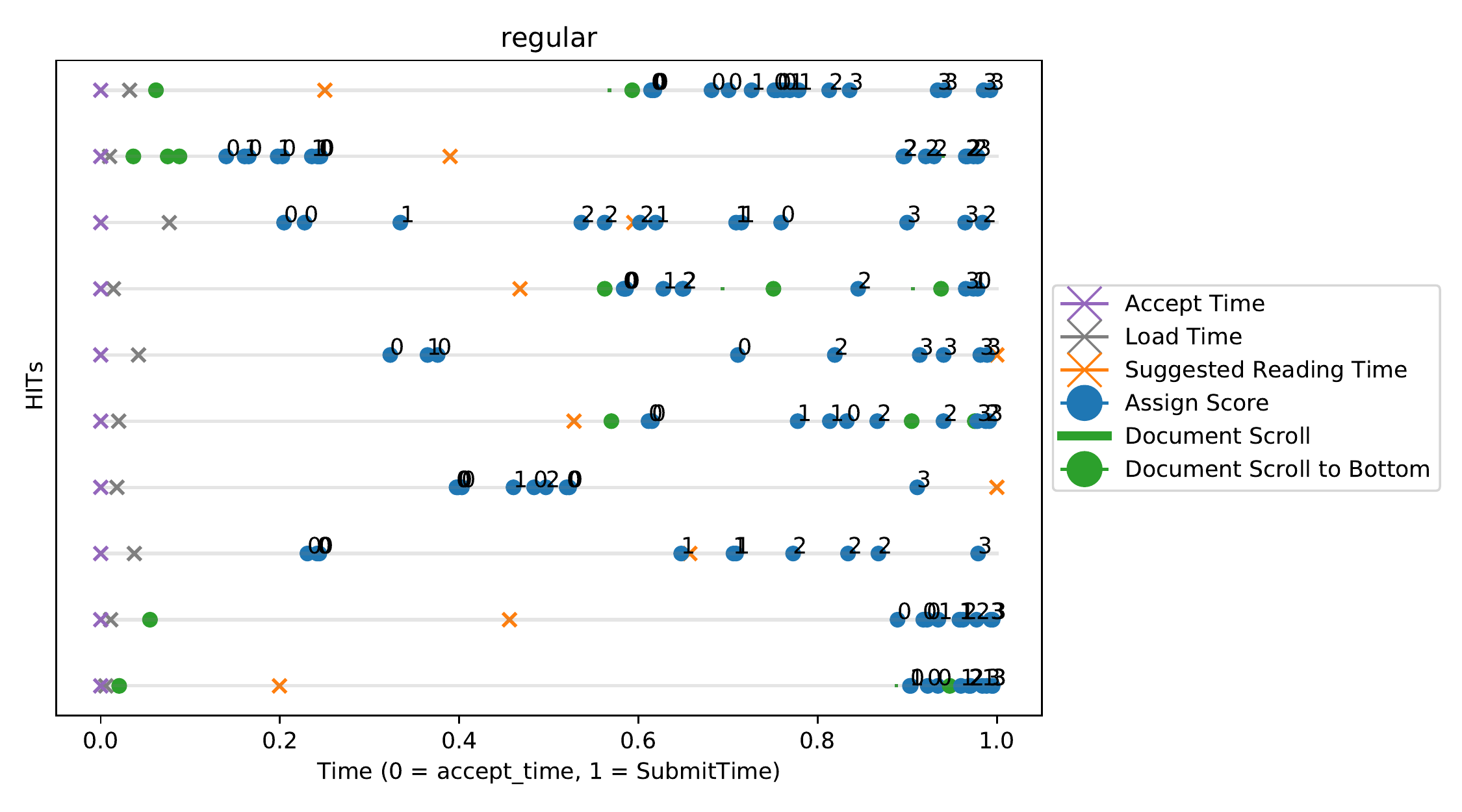}~~~
\includegraphics[width=0.48\textwidth]{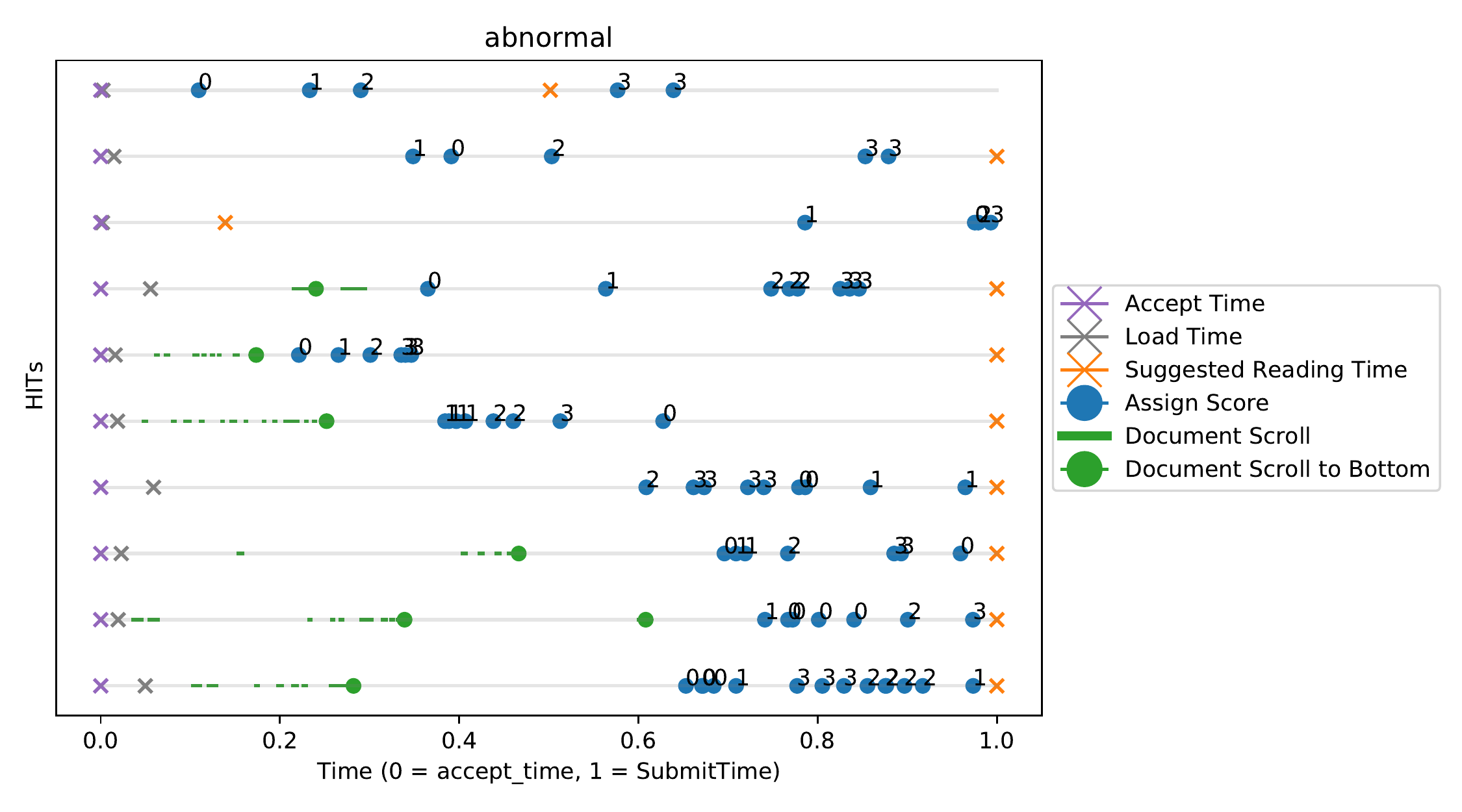}
\end{center}
\caption{Comparison of online behaviors between the abnormal worker ($S_{42}$, right) and the regular worker (left).}
\label{fig:two-worker}
\end{figure*}

\noindent \textbf{Comparison to Expert Ratings} \quad To get an idea of the quality of qualified  MTurk workers according to our approach, we compared their IAA with the IAA obtained by conducting the same endurance task with three researchers as NLG \say{experts}. The pairwise Cohen’s Kappa for all 40 ratings only
%each summary in all HITs has the highest score of 0.387, which is much lower than the scores within or across \textsc{gold} and \textsc{silver} workers. This is also true for the Cohen's Kappa of
reached 0.268 (see Table~\ref{tab:lab} in Appendix~\ref{sec:app-enduranceLab}). The IAA among the experts was comparatively lower than the \textsc{gold} and \textsc{silver} workers, indicating that qualified workers identified by our tasks reached a better agreement at least for the endurance task. Thus, it seems possible to recruit high-quality workers using our pipeline.

\noindent \textbf{Detection of Abnormal Workers} \quad From Cohen's Kappa scores shown in Figure \ref{fig:cat_all}, the worker $S_{42}$\footnote{$S_{42}$ stands for the second \textsc{silver} worker from Round 4} had much lower agreement scores (heatmap in the yellow colors on the row and column corresponding to the worker). Recent studies have uncovered the presence of bots on MTurk~\citep{webb2022too}. To understand the reason for this worker's lower agreement with other workers, we analyzed their online behavior using the metadata extracted from their annotation results.

Figure \ref{fig:two-worker} shows the timeline of each of the 10 HITs as a horizontal gray line. The timelines are plotted from top to bottom, corresponding to the first to the last HIT in the endurance task. The X-axis represents the duration between the time of acceptance and submission, which is normalized by the duration for each HIT (ranging from 0 to 1). Different marks present each annotator behavior, as shown in the legend. Among these behaviors, blue points represent the time when the MTurk worker assigned a score for one of the four summaries, and the corresponding number on top represents the summary index (valued from 0 to 3). Orange crosses denote the suggested reading time of the article in each HIT, given the average human reading speed of 130 words per minute.\footnote{\url{https://wordstotime.com/}} If the suggested reading time after normalization was longer than the duration, we marked the orange cross as 1 at the time of submission which is at the end of the gray line.

Most of the orange crosses were marked at the end of the timelines in Figure \ref{fig:two-worker} (right), indicating this worker assigned scores and submitted the HIT in less time than it usually takes for a human to even finish reading the article. This result demonstrates that this worker may not have put in good faith in the endurance task, which possibly explains the low IAA with other workers. By removing this worker and calculating Krippendorff's Alpha again within \textsc{gold} and \textsc{silver} workers, the IAA increased to 0.454 (compared to 0.396 when including the worker). 
% \sm{Maybe we could present all this the other way around: if this behavior is not found for any other worker; ``from the metadata, we detected that a worker seemed to have an abnormal behavior; this worker is worker XXX, who happens to not correlate as well with the other workers...'' etc. }

% \begin{figure}[ht]
% \begin{center}
% \includegraphics[width=0.5\textwidth]{images/abnormal.png}
% \end{center}
% \caption{Timing Analysis of the Abnormal Worker}
% \label{fig:abnormal}
% \end{figure}

\subsection{Reference-based Task Results}
\label{subsec:refbased-results}
To test the reliability of our qualified workers and compare them to workers who do not undergo our selection process, we launched the reference-based task (see Section \ref{subsec:refbased}), which is open to our qualified workers as well as to any other workers satisfying basic qualification settings.

\noindent \textbf{Qualified Workers after Pipeline} \quad We published the reference-based task to the 12 MTurk workers from four rounds who have passed both the qualification and the endurance task. All 12 workers accepted this task but only 8 workers finished 30 HITs within a week.

\begin{figure}[!h]
\begin{center}
\includegraphics[width=0.35\textwidth]{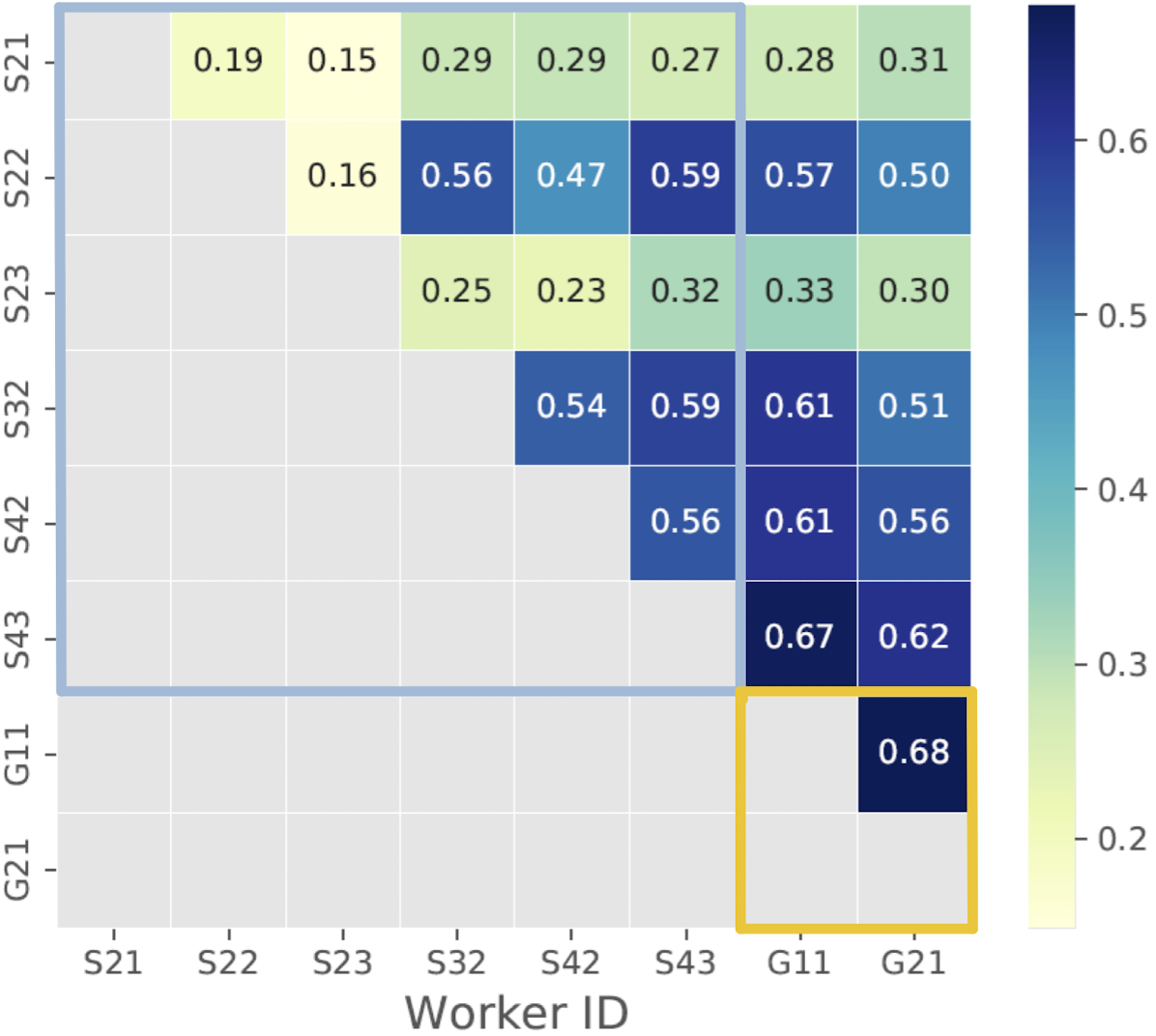}
\end{center}
\caption{Cohen's Kappa for reference-based task (grey frame: \textsc{silver} workers; yellow: \textsc{gold} workers).}
\label{fig:ref}
\end{figure}

There are two scores to evaluate the information coverage between each candidate summary and the reference summary. We use the \say{can2ref} score to represent whether all information in the candidate summary can be found in the reference summary, and the \say{ref2can} score to represent the converse coverage. For both types of scores, we calculated Cohen's Kappa for every worker pair (given 4 candidate summaries per HIT, 30 HITS per worker). Cohen's Kappa for \say{can2ref} score ranges from 0.15 to 0.71, with a relatively high IAA between the first \textsc{gold} workers from the first two rounds ($G_{11}$ and $G_{21}$). Similarly, Cohen's Kappa for \say{ref2can} score ranges from 0.14 to 0.66. Finally, Cohen's Kappa for the combined scores ranges from 0.15 to 0.68 (see Figure \ref{fig:ref}), demonstrating that the agreement numbers are stable across multiple measures. Krippendorff's Alpha for the above scenarios (\say{can2ref} score, \say{ref2can} score, and combined) are 0.558, 0.508, and 0.534.

\noindent \textbf{Baseline MTurk Workers} \quad For comparison, we published the same reference-based task to MTurk workers who did not participate in our previous experiments. 276 MTurk workers participated and each worker finished on average 2 HITs (In total 30 HITs $\times$ 20 Assignments/HIT). Krippendorff’s Alpha for \say{can2ref}, \say{ref2can}, and the two combined were extremely low, at 0.087, 0.077, and 0.080 respectively, demonstrating the necessity of a high-quality recruitment pipeline. We experimented with the following approaches to investigate whether we could increase the agreement between random MTurk workers to a level comparable to qualified workers from our pipeline. 

\begin{description}
   \item[IAA with Median] Among the 20 assignments of each HIT, we randomly divided the workers into 4 groups of 5 workers and took the median of each group representing a \say{new worker} \citep{lau-etal-2014-machine}. Then, we concatenated the results of 20 HITs for the 4 \say{new workers} to calculate IAA. Krippendorff’s Alpha scores increased to 0.191, 0.185, and 0.188 respectively.
   \vspace*{-3mm}
   
   \item[Filter on Timing and Number of Finished HITs] To exclude unqualified workers whose annotations may decrease IAA, only workers who (i) spent more than the suggested reading time\footnote{We performed the same timing analysis as in Section~\ref{subsec:endurance-results}.} and (ii) finished 3 or more HITs were selected for calculation of IAA. This resulted in 25 workers remaining, but Krippendorff’s Alpha remained almost the same as calculated without the filter.
   \vspace*{-3mm}
   
   \item[Statistical Filter (MACE)] We applied the Multi-Annotator Competence Estimation (MACE) \citep{hovy-etal-2013-learning,paun-etal-2018-comparing} to identify reliable workers based on competence scores calculated on annotations. The workers with competence scores above a threshold were kept. We additionally calculated Spearman's coefficient \citep{10.2307/1412159} within the groups of our pipeline and MACE (see Table~\ref{tab:MACE}). We report the results of additional failed attempts to improve Spearman's coefficient across these two groups, in Table \ref{tab:MACE-inter} in the Appendix. 
   \vspace*{-2mm}
\end{description}

\begin{table}[t!]
  \centering
  \resizebox{0.45\textwidth}{!}{
  \begin{tabular}{cccc}
    \toprule
    Threshold & 0.5 & 0.6 & 0.7 \\
    \hline
    \makecell{\% of workers kept} & 19.2\% & 15.9\% & 7.6\% \\
    \makecell{HIT coverage} & 30/30 & 27/30 & 18/30 \\
    \makecell{Avg. num. workers per HIT} & 2.4 & 1.9 & 1.2 \\
    \hline
    \makecell{Krippendorff’s Alpha \\ (all scores)} & 0.380 & 0.472 & 0.754 \\
    \hline
    \makecell{Spearman's coefficient \\ (MACE workers)} & 0.351 & 0.414 & 0.770 \\
    \makecell{Spearman's coefficient \\ (pipeline workers)} & 0.558 & 0.565 & 0.577 \\
    \bottomrule
  \end{tabular}}
  \caption{IAA for different thresholds of MACE.}
  \label{tab:MACE}
\end{table}

\noindent In summary, the most effective methods to improve agreement numbers among random workers were median grouping and MACE. IAA on median scores can raise Krippendorff’s Alpha to almost 0.2. MACE increases Krippendorff’s Alpha as the threshold increases, but at the cost of an incomplete HIT coverage (27/30 and 18/30 respectively for the threshold of 0.6 and 0.7 in Table~\ref{tab:MACE}) and fewer workers per HIT (1.9 and 1.2, respectively, for the threshold of 0.6 and 0.7 in Table~\ref{tab:MACE}). Similarly, Spearman's coefficient of MACE workers can be increased above our pipeline workers' only at the same expense as above.

\noindent \textbf{CloudResearch MTurk Workers} \quad To further test our pipeline, we conducted the same reference-based task on the CloudResearch platform (cloudresearch.com), which helps researchers recruit high-quality annotators. We recruited the same number (eight) of CloudResearch workers as our pipeline. 
The Krippendorff's Alpha and Cohen's Kappa\footnote{The range of Cohen's Kappa is slightly smaller for CloudResearch workers.} for CloudResearch workers is slightly lower than our pipeline workers (see Table \ref{tab:ref-cloudresearch} and Figure \ref{fig:ref-cloudresearch}). Additionally, we found that our pipeline workers have a higher task acceptance rate. This results in a shorter experimental period compared to the task conducted on CloudResearch.

\begin{table}[!h]
\small
\centering
\begin{threeparttable}
  \centering
  \resizebox{0.45\textwidth}{!}{
  \begin{tabular}{c|c|ccc}
    \toprule
    \makecell{Worker\\Source}
    & \makecell{IAA\\Metric}
    & can2ref
    & ref2can
    & \makecell{combined\\score} \\
    \hline
    \multirow{2}{*}{Pipeline}
    & CK & 0.15-0.71 & 0.14-0.66 & 0.15-0.68 \\
    & KA & 0.558 & 0.508 & 0.534 \\
    \hline
    \multirow{2}{*}{\makecell{Cloud\\Research}}
    & CK & 0.18-0.60 & 0.19-0.61 & 0.18-0.60 \\
    & KA & 0.527 & 0.498 & 0.513 \\
    \bottomrule
      \end{tabular}}
  \caption{The range of Cohen's Kappa (CK) and Krippendorff's Alpha (KA) of pipeline and CloudResearch workers for reference-based task.}
  \label{tab:ref-cloudresearch}
  \end{threeparttable}
\end{table}

%In this analysis, we aim to compare the correctness of responses from various annotators and GPT models, namely GPT-3.5, ChatGPT, and GPT-4, against expert judgments.
\noindent \textbf{Analysis of Correctness Across Annotation Sources} \quad We randomly sampled 50 annotation questions from the reference-based task to test correctness, which is defined as the alignment with expert judgments.\footnote{Fifty random samples were chosen in order to differentiate between MACE and pipeline assuming 20\% superiority in terms of correctness.}
In addition, we also compared the expert judgment with scores generated by GPT models: GPT-3.5 (\say{text-davinci-003}) and ChatGPT which are built on InstructGPT \citep{ouyang2022training}, and GPT-4 \citep{OpenAI2023GPT4TR}. Scores are aggregated by taking the median within groups of pipeline, MACE, and CloudResearch workers, as well as experts.\footnote{We use the median of a group of experts as the expert judgment, which has Krippendorff’s Alpha of 0.52.} For ChatGPT we ran inference 5 times with default parameters (temperature=1, top\_p=1) and took the median. To obtain GPT-3.5 and GPT-4 scores temperature was set to 0 with a single run.

\begin{figure}[!h]
\begin{center}
\includegraphics[width=0.45\textwidth]{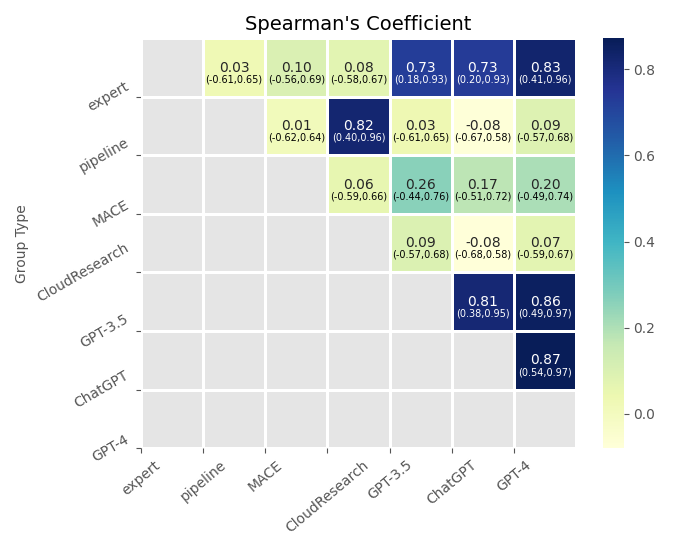}
\end{center}
\caption{Spearman's coefficient for scores of 50 random samples in reference-based task among groups. 95\% confidence interval is shown below the coefficient.}
\label{fig:50-samples-spearsman}
\end{figure}

We did not find that pipeline workers were superior to MACE workers in terms of correctness. Pipeline and CloudResearch workers had a significant Spearman's correlation with each other (see Figure \ref{fig:50-samples-spearsman}), which indicates a reproduction of the recruitment procedure on CloudResearch at a lower cost. However, the confidence intervals are too wide to draw any conclusion about the correlation between crowd annotators and expert judgments (see Table \ref{tab:50-samples-spearsman}). This indicates that the pipeline may not guarantee the training of the correctness of annotations.  However, we found that GPT models correlated well with expert judgments. Further details can be found in Appendix \ref{sec:50-samples} and \ref{sec:appendix-gpt}.
% \footnote{We suspect that there might be a data pollution issue since the dataset used in the reference-based task may be included in the training of GPT models

\begin{table}[!t]
\small
\centering
\begin{threeparttable}
  \centering
  \resizebox{0.45\textwidth}{!}{
  \begin{tabular}{cccc}
    \toprule
    Class
    & Group Type
    & \makecell{Spearman's\\Coefficient}
    & \makecell{95\% Confidence\\Interval} \\
    \hline
    \multirow{3}{*}{\makecell{Crowd\\Annotators}}
    & Pipeline & 0.03 & (-0.61, 0.65) \\
    & MACE & 0.10 & (-0.56, 0.69) \\
    & CloudResearch & 0.08 & (-0.58, 0.67) \\
    \hline
    \multirow{3}{*}{\makecell{GPT\\models}}
    & GPT-3.5 & 0.73 & (0.18, 0.93) \\
    & ChatGPT & 0.73 & (0.20, 0.93) \\
    & GPT-4 & 0.83 & (0.41, 0.96) \\
    \bottomrule
  \end{tabular}}
  \caption{Spearman's coefficient of the expert judgment and groups for crowd annotators and GPT models.}
  \label{tab:50-samples-spearsman}
  \end{threeparttable}
\end{table}

\subsection{Discussion}
\label{subsec:discussion}
% In Section~\ref{subsec:endurance-results}, we show that our 12 qualified workers carried out the endurance task with a higher IAA than our expert lab annotators. In Section~\ref{subsec:refbased-results}, out of the 12 \textsc{gold} and \textsc{silver} annotators, 8 carried out the subsequent reference-based task, which we also ran with non-expert annotators. We increase the IAA of the non-expert annotators using MACE filtering. 

% The first MACE threshold that provided full coverage of the HITs, and was thus comparable to our pipeline, was 0.5; for this threshold, 19.2\% of the workers were kept, and we obtained 2.4 annotations per HIT for a Krippendorf's Alpha of 0.380. Even though the pipeline approach kept only 4\% of the initial crowd performing the reference-based task, this yielded more than 3 times the number of annotations for each HIT, and reached an average Krippendorf's Alpha of 0.534 on all worker pairs; see summary in  Table~\ref{tab:MACE-compar}. Furthermore, the cost for the average number of workers per HIT using MACE was significantly higher than those acquired using our pipeline. 

% According to our experiment, we not only obtained reliable annotators who provided high-quality evaluations but did so while tremendously reducing the manual workload, since we filtered workers \textit{before} they took part in the actual evaluation task. By comparison, with MACE more than 80\% of the annotations have to be discarded \textit{after} carrying out the task, which means a waste of time and resources. 

In Section~\ref{subsec:refbased-results}, we published the same reference-based task as a test to different crowd annotators (pipeline, MACE, and CloudResearch). It showed that filtering workers \textit{before} the actual evaluation task (pipeline) can avoid the waste of time and resources and achieve high agreement at a lower cost and a full coverage of HITs, compared to discarding annotations \textit{after} the task (MACE) (see Table \ref{tab:MACE-compar}). Our pipeline also recruited workers of similar quality to CloudResearch at a lower cost; however, based on further analysis, the correctness of annotations was not guaranteed (see Section \ref{sec:limitation} for details). Besides, details about the estimated cost of GPT models for the reference-based task can be found in Table \ref{tab:gpt-cost} in Appendix \ref{sec:app-gpt-cost}.

% TABLE THAT COMPARES APPROACHES MACE
\begin{table}[h!]
  \centering
  \resizebox{0.48\textwidth}{!}{
  \begin{tabular}{cccc}
    \toprule
    & Pipeline & MACE (0.5) & CloudResearch \\
    \hline
    Num. of initial workers & 200 & 276 & 45 \\
    \% of workers kept & 4\% & 19.2\% & 17.8\% \\
    HIT coverage & 30/30 & 30/30 & 30/30 \\
    \hline
    \makecell{Avg. num. workers per HIT} & 8 & 2.4 & 8 \\
    Krippendorff’s Alpha & \textbf{0.534} & 0.380 & 0.513 \\
    \hline
    \makecell{Cost per worker\\(for Avg. num. workers per HIT)} & \textbf{\$27} & \$175 & \$31 \\
    \bottomrule
  \end{tabular}}
  \caption{Comparison between approaches of crowd annotators (pipeline, MACE, and CloudResearch) for the reference-based task.}
  \label{tab:MACE-compar}
\end{table}

\section{Statistical Test for Stability of Pipeline}
% We notice the numbers of MTurk workers who have passed the qualification task or the endurance task for each round are similar.
We next examined whether there was a difference in the probability of passing the qualification and endurance task among MTurk workers. Thus, we started by assuming the probability of passing each task for each round came from the same distribution, and we performed a statistical test as follows.

\begin{table}[b!]
  \centering
  \resizebox{0.4\textwidth}{!}{
  \begin{tabular}{ccc}
    \toprule
    Annotation Task & Qual. Task & End. Task \\
    \hline
    Pass Rate & 0.13 & 0.06 \\
    \hline
    \makecell{Mean of \\ Pass Rate (Bootstrap)} & 0.1302 & 0.0602 \\
    \hline
    \makecell{Standard Dev. of \\ Pass Rate (Bootstrap)} & 0.0236& 0.0168 \\
    \bottomrule
  \end{tabular}}
    \caption{Statistical test results for stability of pipeline.}
  \label{tab:stat}
\end{table}

Let $\mathcal{X}$ denote the random variable representing the MTurk worker. For the qualification task, let $q_{x\in\mathcal{X}}(x)$ denote the binary random variable which has the value of 1 if the worker can pass the task, and 0 otherwise. Similarly, let $e_{x\in\mathcal{X}}(x)$ denote the binary random variable indicating whether the worker can pass the endurance task. Given 50 MTurk workers in each round, we use $Q$ to denote the binary random variables in a round as (\ref{eq:Q}). It can also be regarded as examples sampled from $q_{x\in\mathcal{X}}(x)$. Among the samples, the probability of a worker who can pass the qualification task is equal to the expectation of $q_{x\in\mathcal{X}}(x)=1$ as (\ref{eq:p_q=1}). Since only workers who passed the qualification task are eligible for the endurance task, the probability of a worker passing the endurance task is equal to the expectation of $e_{x\in\mathcal{X}, q(x)=1}(x)=1$ as (\ref{eq:p_e=1}), which is a joint distribution of $q_{x\in\mathcal{X}}(x)$ and $e_{x\in\mathcal{X}}(x)$.

\vspace*{-7mm}
\begin{equation}
    \label{eq:Q}
    Q = \{q_{x_1\in\mathcal{X}}(x_1), ..., q_{x_{50}\in\mathcal{X}}(x_{50})\}
\end{equation}

\vspace*{-10mm}
\begin{equation}
    \label{eq:p_q=1}
    P(q_{x\in\mathcal{X}}(x)=1) = \mathbb{E}(q_{x\in\mathcal{X}}(x)=1)
\end{equation}

\vspace*{-10mm}
\begin{align}
    \label{eq:p_e=1}
    &P(e_{x\in\mathcal{X}, q(x)=1}(x)=1) \notag \\
    = &\mathbb{E}(e_{x\in\mathcal{X}, q(x)=1}(x)=1) \\
    = &P(e_{x\in\mathcal{X}}(x)=1|q(x)=1) P(q(x)=1) \notag 
\end{align}

\vspace*{-2mm}
Thus, we used the Bootstrap method \citep{efron1992bootstrap} with 10,000 iterations to estimate the mean and standard deviation of the probability of passing the qualification and endurance task. Table \ref{tab:stat} shows the results of all rounds with breakdowns of each round. We can see some variance that might come from MTurk workers given each round. To test whether there is a difference in the probability of passing each task among different rounds, we conducted the permutation test \citep{fisher1936design,10.2307/2984124} for every two rounds. The results show that we cannot reject the null hypothesis that the underlying distributions of every two rounds are the same (see Appendix \ref{sec:app-statTestRound}).

\section{Conclusion}
In this paper, we present a two-step recruitment pipeline that yields 12 qualified workers (4 \textsc{gold} and 8 \textsc{silver} workers) out of 200 MTurk workers with basic qualification settings in our experiments. 
We show that workers identified by our pipeline can (i) achieve a higher inter-annotator agreement than expert annotators in the endurance task, (ii) outperform the statistical filter (MACE) that discards annotation \textit{after} the reference-based task, and (iii) replicate a proxy of CloudResearch annotations in the correctness analysis.
Though the 6\% yield rate is not as expected, our pipeline serves as the \textbf{best practice} to deliver high-agreement annotations and addresses the widespread waste of resources on low-quality annotations through filtering out subpar workers \textit{before} they embark on large-scale tasks.
In the future, we plan to build up a pool of reliable annotators who can deliver high-quality (both high agreement and correctness) evaluations on a large scale and in multiple tasks, languages, and platforms.

\clearpage
\section{Limitations}
\label{sec:limitation}
This research creates a relatively complete pipeline to identify qualified MTurk workers for high-quality human evaluations based on existing techniques, and thoroughly tests the effectiveness of this pipeline both qualitatively and quantitatively compared to other methods. However, there are several limitations of this work:
\vspace*{-3mm}
\begin{itemize}[leftmargin=0.7cm]
  \item \textbf{The experiments are only conducted for summarization tasks in English on MTurk platform.} Thus, this pipeline can also be tested on other NLG tasks, in other languages, and on other platforms to see whether our three-step concept generalizes broadly to all human evaluations.
  \vspace*{-3mm}
  \item \textbf{The specific questions designed for each task are not \say{panacea} solutions.} A better HIT design may exist for different experimental purposes, as long as it follows the ideas behind each task. For example, the endurance task aims to ensure the worker's reliable performance on a large number of annotations, so modifications based on this idea might work better in case-by-case scenarios\footnote{We encourage starting the design from the reference-based task (which performs as the test of true annotation task) and thinking about what specific training the annotators are expected to have through the qualification and endurance task.}.
  \vspace*{-3mm}
  \item \textbf{There is no guarantee for the training of correctness in the pipeline though a high agreement is achieved.} An additional correctness check might need to be included along with the endurance task to achieve both high agreement and correctness through the filtering of the pipeline.
\end{itemize}

\section{Ethical Considerations}
Considering that crowd workers are often underpaid, experiments in this work all followed fair working wage standards\footnote{\url{https://livingwage.mit.edu/counties/27053}} when using MTurk for recruitment purposes (details for each task are in Table \ref{tab:wage-table}). In addition, we have not rejected the work from any unqualified workers so far, though we reserve the right to do so when conducting the experiments.

In our experiments, personal data (any information relating to an identifiable natural person) was collected, processed, and stored based on certain data protection regulations,\footnote{\url{https://gdpr.eu/article-4-definitions/}} given relevant privacy concerns. Special category information (i.e. personal data revealing racial or ethnic origin, etc.) was not included in this work. More information about the details of human evaluation experiments in this work can be found in the Human Evaluation Datasheet (HEDS) \citep{shimorina-belz-2022-human} in the Appendix.

\section*{Acknowledgements}
We would like to thank the anonymous reviewers for their helpful feedback on our paper. We would like to thank Claire Daniele for her editorial support. Mille's contribution was funded by the European Union under the Marie Skłodowska-Curie grant agreement No 101062572 (M-FleNS).

\bibliography{ref}

\begin{thebibliography}{44}
\expandafter\ifx\csname natexlab\endcsname\relax\def\natexlab#1{#1}\fi

\bibitem[{Amidei et~al.(2020)Amidei, Piwek, and
  Willis}]{amidei-etal-2020-identifying}
Jacopo Amidei, Paul Piwek, and Alistair Willis. 2020.
\newblock \href {https://doi.org/10.18653/v1/2020.coling-main.421} {Identifying
  annotator bias: A new {IRT}-based method for bias identification}.
\newblock In \emph{Proceedings of the 28th International Conference on
  Computational Linguistics}, pages 4787--4797, Barcelona, Spain (Online).
  International Committee on Computational Linguistics.

\bibitem[{Arechar et~al.(2017)Arechar, Kraft-Todd, and
  Rand}]{arechar2017turking}
Antonio~A Arechar, Gordon~T Kraft-Todd, and David~G Rand. 2017.
\newblock Turking overtime: How participant characteristics and behavior vary
  over time and day on amazon mechanical turk.
\newblock \emph{Journal of the Economic Science Association}, 3(1):1--11.

\bibitem[{Banerjee and Lavie(2005)}]{banerjee-lavie-2005-meteor}
Satanjeev Banerjee and Alon Lavie. 2005.
\newblock \href {https://aclanthology.org/W05-0909} {{METEOR}: An automatic
  metric for {MT} evaluation with improved correlation with human judgments}.
\newblock In \emph{Proceedings of the {ACL} Workshop on Intrinsic and Extrinsic
  Evaluation Measures for Machine Translation and/or Summarization}, pages
  65--72, Ann Arbor, Michigan. Association for Computational Linguistics.

\bibitem[{Berinsky et~al.(2012)Berinsky, Huber, and
  Lenz}]{berinsky_huber_lenz_2012}
Adam~J. Berinsky, Gregory~A. Huber, and Gabriel~S. Lenz. 2012.
\newblock \href {https://doi.org/10.1093/pan/mpr057} {Evaluating online labor
  markets for experimental research: Amazon.com's mechanical turk}.
\newblock \emph{Political Analysis}, 20(3):351–368.

\bibitem[{Buhrmester et~al.(2011)Buhrmester, Kwang, and
  Gosling}]{buhrmester2016amazon}
Michael Buhrmester, Tracy Kwang, and Samuel~D Gosling. 2011.
\newblock Amazon's mechanical turk: A new source of inexpensive, yet
  high-quality data?
\newblock \emph{Perspectives on Psychological Science}, 6(1):3--5.

\bibitem[{Callison-Burch(2009)}]{callison-burch-2009-fast}
Chris Callison-Burch. 2009.
\newblock \href {https://aclanthology.org/D09-1030} {Fast, cheap, and creative:
  Evaluating translation quality using {A}mazon{'}s {M}echanical {T}urk}.
\newblock In \emph{Proceedings of the 2009 Conference on Empirical Methods in
  Natural Language Processing}, pages 286--295, Singapore. Association for
  Computational Linguistics.

\bibitem[{Cohen(1960)}]{Cohen1960ACO}
Jacob Cohen. 1960.
\newblock A coefficient of agreement for nominal scales.
\newblock \emph{Educational and Psychological Measurement}, 20:37 -- 46.

\bibitem[{Devlin et~al.(2019)Devlin, Chang, Lee, and
  Toutanova}]{devlin-etal-2019-bert}
Jacob Devlin, Ming-Wei Chang, Kenton Lee, and Kristina Toutanova. 2019.
\newblock \href {https://doi.org/10.18653/v1/N19-1423} {{BERT}: Pre-training of
  deep bidirectional transformers for language understanding}.
\newblock In \emph{Proceedings of the 2019 Conference of the North {A}merican
  Chapter of the Association for Computational Linguistics: Human Language
  Technologies, Volume 1 (Long and Short Papers)}, pages 4171--4186,
  Minneapolis, Minnesota. Association for Computational Linguistics.

\bibitem[{Efron(1992)}]{efron1992bootstrap}
Bradley Efron. 1992.
\newblock Bootstrap methods: another look at the jackknife.
\newblock In \emph{Breakthroughs in statistics}, pages 569--593. Springer.

\bibitem[{Fabbri et~al.(2021)Fabbri, Kry{\'s}ci{\'n}ski, McCann, Xiong, Socher,
  and Radev}]{fabbri-etal-2021-summeval}
Alexander~R. Fabbri, Wojciech Kry{\'s}ci{\'n}ski, Bryan McCann, Caiming Xiong,
  Richard Socher, and Dragomir Radev. 2021.
\newblock \href {https://doi.org/10.1162/tacl_a_00373} {{S}umm{E}val:
  Re-evaluating summarization evaluation}.
\newblock \emph{Transactions of the Association for Computational Linguistics},
  9:391--409.

\bibitem[{Fisher(1936)}]{fisher1936design}
Ronald~Aylmer Fisher. 1936.
\newblock Design of experiments.
\newblock \emph{British Medical Journal}, 1(3923):554.

\bibitem[{Gehrmann et~al.(2022)Gehrmann, Clark, and
  Sellam}]{gehrmann2022repairing}
Sebastian Gehrmann, Elizabeth Clark, and Thibault Sellam. 2022.
\newblock Repairing the cracked foundation: A survey of obstacles in evaluation
  practices for generated text.
\newblock \emph{arXiv preprint arXiv:2202.06935}.

\bibitem[{Gillick and Liu(2010)}]{gillick-liu-2010-non}
Dan Gillick and Yang Liu. 2010.
\newblock \href {https://aclanthology.org/W10-0722} {Non-expert evaluation of
  summarization systems is risky}.
\newblock In \emph{Proceedings of the {NAACL} {HLT} 2010 Workshop on Creating
  Speech and Language Data with {A}mazon{'}s Mechanical Turk}, pages 148--151,
  Los Angeles. Association for Computational Linguistics.

\bibitem[{Graham et~al.(2018)Graham, Awad, and
  Smeaton}]{10.1371/journal.pone.0202789}
Yvette Graham, George Awad, and Alan Smeaton. 2018.
\newblock \href {https://doi.org/10.1371/journal.pone.0202789} {Evaluation of
  automatic video captioning using direct assessment}.
\newblock \emph{PLOS ONE}, 13(9):1--20.

\bibitem[{Graham et~al.(2017)Graham, Baldwin, Moffat, and
  Zobel}]{graham_baldwin_moffat_zobel_2017}
Yvette Graham, Timothy Baldwin, Alistair Moffat, and Justin Zobel. 2017.
\newblock Can machine translation systems be evaluated by the crowd alone.
\newblock \emph{Natural Language Engineering}, 23(1):3--30.

\bibitem[{Hardy et~al.(2019)Hardy, Narayan, and
  Vlachos}]{hardy-etal-2019-highres}
Hardy Hardy, Shashi Narayan, and Andreas Vlachos. 2019.
\newblock \href {https://doi.org/10.18653/v1/P19-1330} {{H}igh{RES}:
  Highlight-based reference-less evaluation of summarization}.
\newblock In \emph{Proceedings of the 57th Annual Meeting of the Association
  for Computational Linguistics}, pages 3381--3392, Florence, Italy.
  Association for Computational Linguistics.

\bibitem[{Hayes and Krippendorff(2007)}]{hayes2007answering}
Andrew~F Hayes and Klaus Krippendorff. 2007.
\newblock Answering the call for a standard reliability measure for coding
  data.
\newblock \emph{Communication methods and measures}, 1(1):77--89.

\bibitem[{Hovy et~al.(2013)Hovy, Berg-Kirkpatrick, Vaswani, and
  Hovy}]{hovy-etal-2013-learning}
Dirk Hovy, Taylor Berg-Kirkpatrick, Ashish Vaswani, and Eduard Hovy. 2013.
\newblock \href {https://aclanthology.org/N13-1132} {Learning whom to trust
  with {MACE}}.
\newblock In \emph{Proceedings of the 2013 Conference of the North {A}merican
  Chapter of the Association for Computational Linguistics: Human Language
  Technologies}, pages 1120--1130, Atlanta, Georgia. Association for
  Computational Linguistics.

\bibitem[{Howcroft et~al.(2020)Howcroft, Belz, Clinciu, Gkatzia, Hasan,
  Mahamood, Mille, van Miltenburg, Santhanam, and
  Rieser}]{howcroft-etal-2020-twenty}
David~M. Howcroft, Anya Belz, Miruna-Adriana Clinciu, Dimitra Gkatzia, Sadid~A.
  Hasan, Saad Mahamood, Simon Mille, Emiel van Miltenburg, Sashank Santhanam,
  and Verena Rieser. 2020.
\newblock \href {https://aclanthology.org/2020.inlg-1.23} {Twenty years of
  confusion in human evaluation: {NLG} needs evaluation sheets and standardised
  definitions}.
\newblock In \emph{Proceedings of the 13th International Conference on Natural
  Language Generation}, pages 169--182, Dublin, Ireland. Association for
  Computational Linguistics.

\bibitem[{Huynh et~al.(2021)Huynh, Bigham, and Eskenazi}]{huynh2021survey}
Jessica Huynh, Jeffrey Bigham, and Maxine Eskenazi. 2021.
\newblock A survey of nlp-related crowdsourcing hits: what works and what does
  not.
\newblock \emph{arXiv preprint arXiv:2111.05241}.

\bibitem[{Isozaki et~al.(2010)Isozaki, Hirao, Duh, Sudoh, and
  Tsukada}]{isozaki-etal-2010-automatic}
Hideki Isozaki, Tsutomu Hirao, Kevin Duh, Katsuhito Sudoh, and Hajime Tsukada.
  2010.
\newblock \href {https://aclanthology.org/D10-1092} {Automatic evaluation of
  translation quality for distant language pairs}.
\newblock In \emph{Proceedings of the 2010 Conference on Empirical Methods in
  Natural Language Processing}, pages 944--952, Cambridge, MA. Association for
  Computational Linguistics.

\bibitem[{Kaplan and Meier(1958)}]{km-estimator}
E.~L. Kaplan and Paul Meier. 1958.
\newblock \href {http://www.jstor.org/stable/2281868} {Nonparametric estimation
  from incomplete observations}.
\newblock \emph{Journal of the American Statistical Association},
  53(282):457--481.

\bibitem[{Karpinska et~al.(2021)Karpinska, Akoury, and
  Iyyer}]{karpinska-etal-2021-perils}
Marzena Karpinska, Nader Akoury, and Mohit Iyyer. 2021.
\newblock \href {https://doi.org/10.18653/v1/2021.emnlp-main.97} {The perils of
  using {M}echanical {T}urk to evaluate open-ended text generation}.
\newblock In \emph{Proceedings of the 2021 Conference on Empirical Methods in
  Natural Language Processing}, pages 1265--1285, Online and Punta Cana,
  Dominican Republic. Association for Computational Linguistics.

\bibitem[{Kummerfeld(2021)}]{kummerfeld2021quantifying}
Jonathan~K. Kummerfeld. 2021.
\newblock \href {https://doi.org/10.18653/v1/2021.acl-short.44} {Quantifying
  and avoiding unfair qualification labour in crowdsourcing}.
\newblock In \emph{Proceedings of the 59th Annual Meeting of the Association
  for Computational Linguistics and the 11th International Joint Conference on
  Natural Language Processing (Volume 2: Short Papers)}, pages 343--349,
  Online. Association for Computational Linguistics.

\bibitem[{Lau et~al.(2014)Lau, Newman, and Baldwin}]{lau-etal-2014-machine}
Jey~Han Lau, David Newman, and Timothy Baldwin. 2014.
\newblock \href {https://doi.org/10.3115/v1/E14-1056} {Machine reading tea
  leaves: Automatically evaluating topic coherence and topic model quality}.
\newblock In \emph{Proceedings of the 14th Conference of the {E}uropean Chapter
  of the Association for Computational Linguistics}, pages 530--539,
  Gothenburg, Sweden. Association for Computational Linguistics.

\bibitem[{Likert(1932)}]{likert1932technique}
Rensis Likert. 1932.
\newblock A technique for the measurement of attitudes.
\newblock \emph{Archives of psychology}.

\bibitem[{Lin(2004)}]{lin-2004-rouge}
Chin-Yew Lin. 2004.
\newblock \href {https://aclanthology.org/W04-1013} {{ROUGE}: A package for
  automatic evaluation of summaries}.
\newblock In \emph{Text Summarization Branches Out}, pages 74--81, Barcelona,
  Spain. Association for Computational Linguistics.

\bibitem[{Manning et~al.(2020)Manning, Wein, and
  Schneider}]{manning-etal-2020-human}
Emma Manning, Shira Wein, and Nathan Schneider. 2020.
\newblock \href {https://doi.org/10.18653/v1/2020.coling-main.420} {A human
  evaluation of {AMR}-to-{E}nglish generation systems}.
\newblock In \emph{Proceedings of the 28th International Conference on
  Computational Linguistics}, pages 4773--4786, Barcelona, Spain (Online).
  International Committee on Computational Linguistics.

\bibitem[{Mehri et~al.(2022)Mehri, Choi, D’Haro, Deriu, Esk{\'e}nazi, Gasic,
  Georgila, Hakkani-T{\"u}r, Li, Rieser, Shaikh, Traum, Yeh, Yu, Zhang, and
  Zhang}]{Mehri2022ReportFT}
Shikib Mehri, Jinho Choi, L.~F. D’Haro, Jan Deriu, Maxine Esk{\'e}nazi,
  Milica Gasic, Kallirroi Georgila, Dilek~Z. Hakkani-T{\"u}r, Zekang Li, Verena
  Rieser, Samira Shaikh, David~R. Traum, Yi-Ting Yeh, Zhou Yu, Yizhe Zhang, and
  Chen Zhang. 2022.
\newblock Report from the nsf future directions workshop on automatic
  evaluation of dialog: Research directions and challenges.
\newblock \emph{ArXiv}, abs/2203.10012.

\bibitem[{Mille et~al.(2019)Mille, Belz, Bohnet, Graham, and
  Wanner}]{mille-etal-2019-second}
Simon Mille, Anja Belz, Bernd Bohnet, Yvette Graham, and Leo Wanner. 2019.
\newblock \href {https://doi.org/10.18653/v1/D19-6301} {The second multilingual
  surface realisation shared task ({SR}{'}19): Overview and evaluation
  results}.
\newblock In \emph{Proceedings of the 2nd Workshop on Multilingual Surface
  Realisation (MSR 2019)}, pages 1--17, Hong Kong, China. Association for
  Computational Linguistics.

\bibitem[{OpenAI(2023)}]{OpenAI2023GPT4TR}
OpenAI. 2023.
\newblock Gpt-4 technical report.
\newblock \emph{ArXiv}, abs/2303.08774.

\bibitem[{Oppenheimer et~al.(2009)Oppenheimer, Meyvis, and
  Davidenko}]{OPPENHEIMER2009867}
Daniel~M. Oppenheimer, Tom Meyvis, and Nicolas Davidenko. 2009.
\newblock \href {https://doi.org/https://doi.org/10.1016/j.jesp.2009.03.009}
  {Instructional manipulation checks: Detecting satisficing to increase
  statistical power}.
\newblock \emph{Journal of Experimental Social Psychology}, 45(4):867--872.

\bibitem[{Oppenlaender et~al.(2020)Oppenlaender, Milland, Visuri, Ipeirotis,
  and Hosio}]{Oppenlaender2020}
Jonas Oppenlaender, Kristy Milland, Aku Visuri, Panos Ipeirotis, and Simo
  Hosio. 2020.
\newblock \href {https://doi.org/10.1145/3313831.3376677} {Creativity on paid
  crowdsourcing platforms}.
\newblock In \emph{Proceedings of the 2020 CHI Conference on Human Factors in
  Computing Systems}, CHI '20, page 1–14, New York, NY, USA. Association for
  Computing Machinery.

\bibitem[{Ouyang et~al.(2022)Ouyang, Wu, Jiang, Almeida, Wainwright, Mishkin,
  Zhang, Agarwal, Slama, Ray et~al.}]{ouyang2022training}
Long Ouyang, Jeffrey Wu, Xu~Jiang, Diogo Almeida, Carroll Wainwright, Pamela
  Mishkin, Chong Zhang, Sandhini Agarwal, Katarina Slama, Alex Ray, et~al.
  2022.
\newblock Training language models to follow instructions with human feedback.
\newblock \emph{Advances in Neural Information Processing Systems},
  35:27730--27744.

\bibitem[{Paolacci et~al.(2010)Paolacci, Chandler, and
  Ipeirotis}]{paolacci2010running}
Gabriele Paolacci, Jesse Chandler, and Panagiotis~G Ipeirotis. 2010.
\newblock Running experiments on amazon mechanical turk.
\newblock \emph{Judgment and Decision making}, 5(5):411--419.

\bibitem[{Papineni et~al.(2002)Papineni, Roukos, Ward, and
  Zhu}]{10.3115/1073083.1073135}
Kishore Papineni, Salim Roukos, Todd Ward, and Wei-Jing Zhu. 2002.
\newblock \href {https://doi.org/10.3115/1073083.1073135} {Bleu: A method for
  automatic evaluation of machine translation}.
\newblock In \emph{Proceedings of the 40th Annual Meeting on Association for
  Computational Linguistics}, ACL '02, page 311–318, USA. Association for
  Computational Linguistics.

\bibitem[{Paun et~al.(2018)Paun, Carpenter, Chamberlain, Hovy, Kruschwitz, and
  Poesio}]{paun-etal-2018-comparing}
Silviu Paun, Bob Carpenter, Jon Chamberlain, Dirk Hovy, Udo Kruschwitz, and
  Massimo Poesio. 2018.
\newblock \href {https://doi.org/10.1162/tacl_a_00040} {Comparing {B}ayesian
  models of annotation}.
\newblock \emph{Transactions of the Association for Computational Linguistics},
  6:571--585.

\bibitem[{Pitman(1937)}]{10.2307/2984124}
E.~J.~G. Pitman. 1937.
\newblock \href {http://www.jstor.org/stable/2984124} {Significance tests which
  may be applied to samples from any populations}.
\newblock \emph{Supplement to the Journal of the Royal Statistical Society},
  4(1):119--130.

\bibitem[{Robinson et~al.(2019)Robinson, Rosenzweig, Moss, and
  Litman}]{10.1371/journal.pone.0226394}
Jonathan Robinson, Cheskie Rosenzweig, Aaron~J. Moss, and Leib Litman. 2019.
\newblock \href {https://doi.org/10.1371/journal.pone.0226394} {Tapped out or
  barely tapped? recommendations for how to harness the vast and largely unused
  potential of the mechanical turk participant pool}.
\newblock \emph{PLOS ONE}, 14(12):1--29.

\bibitem[{Shimorina and Belz(2022)}]{shimorina-belz-2022-human}
Anastasia Shimorina and Anya Belz. 2022.
\newblock \href {https://aclanthology.org/2022.humeval-1.6} {The human
  evaluation datasheet: A template for recording details of human evaluation
  experiments in {NLP}}.
\newblock In \emph{Proceedings of the 2nd Workshop on Human Evaluation of NLP
  Systems (HumEval)}, pages 54--75, Dublin, Ireland. Association for
  Computational Linguistics.

\bibitem[{Spearman(1904)}]{10.2307/1412159}
C.~Spearman. 1904.
\newblock \href {http://www.jstor.org/stable/1412159} {The proof and
  measurement of association between two things}.
\newblock \emph{The American Journal of Psychology}, 15(1):72--101.

\bibitem[{Webb and Tangney(2022)}]{webb2022too}
Margaret~A Webb and June~P Tangney. 2022.
\newblock Too good to be true: Bots and bad data from mechanical turk.
\newblock \emph{Perspectives on Psychological Science}, page 17456916221120027.

\bibitem[{Whiting et~al.(2019)Whiting, Hugh, and
  Bernstein}]{Whiting_Hugh_Bernstein_2019}
Mark~E. Whiting, Grant Hugh, and Michael~S. Bernstein. 2019.
\newblock \href {https://doi.org/10.1609/hcomp.v7i1.5283} {Fair work: Crowd
  work minimum wage with one line of code}.
\newblock \emph{Proceedings of the AAAI Conference on Human Computation and
  Crowdsourcing}, 7(1):197--206.

\bibitem[{Zhang et~al.(2020)Zhang, Kishore, Wu, Weinberger, and
  Artzi}]{Zhang*2020BERTScore:}
Tianyi Zhang, Varsha Kishore, Felix Wu, Kilian~Q. Weinberger, and Yoav Artzi.
  2020.
\newblock \href {https://openreview.net/forum?id=SkeHuCVFDr} {Bertscore:
  Evaluating text generation with bert}.
\newblock In \emph{International Conference on Learning Representations}.

\end{thebibliography}
\bibliographystyle{acl_natbib}

%%%%%%%%%%%%%%%%%%%%%%%%%%%%%%%%%%%%%%%%%%%%%%%%%%%%%%%%%
\clearpage
\onecolumn
\appendix

\section{Appendix}

\subsection{Proportion of Worker Categories in Qualification Task for Each Round}
\begin{table*}[!bht]
% \small
\centering
\begin{threeparttable}
  \centering
  \begin{tabular}{ccccccc}
    \toprule
    \multicolumn{2}{c}{Annotation Task} 
    & \makecell{Total Number \\ of Workers}
    & \makecell{\textsc{gold} \\ Workers}
    & \makecell{\textsc{silver} \\ Workers}
    & \makecell{\textsc{broze} \\ Workers}
    & \makecell{\textsc{block} \\ Workers} \\
    \hline
    \multirow{4}{*}{\makecell{Qualification \\ Task}}
    & Round 1 & 50 & 1 (2\%) & 4 (8\%) & 32 (64\%) & 13 (26\%) \\
    & Round 2 & 50 & 3 (6\%) & 5 (10\%) & 29 (58\%) & 13 (26\%) \\
    & Round 3 & 50 & 2 (4\%) & 3 (6\%) & 24 (48\%) & 21 (42\%) \\
    & Round 4 & 50 & 2 (4\%) & 6 (12\%) & 27 (54\%) & 15 (30\%) \\
    \bottomrule
  \end{tabular}
  \caption{Proportion of worker categories for each round.}
  \label{tab:type-table}
  \end{threeparttable}
\end{table*}

\subsection{Cohen’s Kappa for Each Summary in Endurance Task}
\label{sec:app-kappaEndurance}
For the figures below, \say{Answer.score\_0} to \say{Answer.score\_3} correspond to the scores aggregated from the 1st to 4th summary separately for each HIT. The dark color indicates a high IAA in terms of Cohen's Kappa score. $S42$ stands for the second \textsc{silver} worker from Round 4.

\begin{figure*}[!ht]
\begin{center}
\includegraphics[width=1\textwidth]{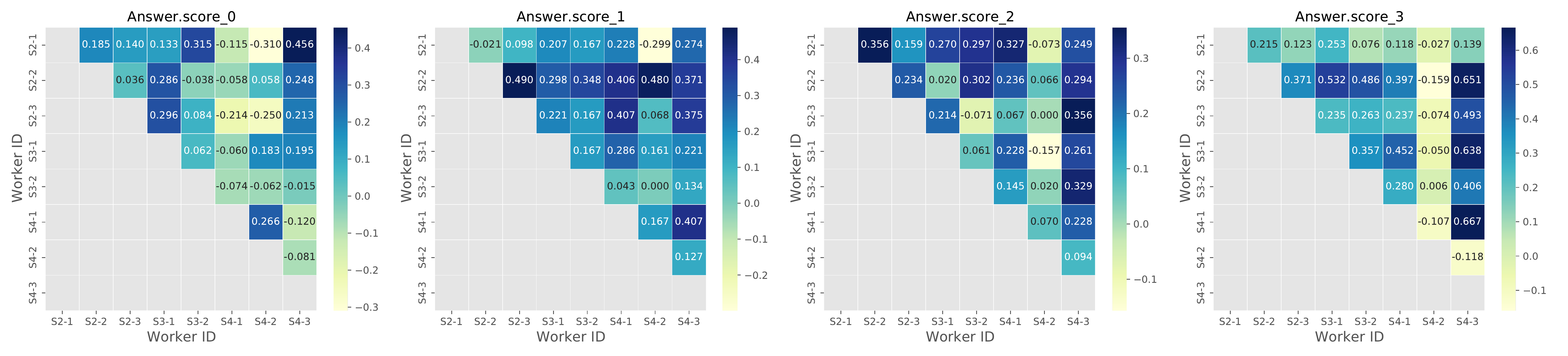}
\end{center}
\vspace{-0.2cm}
\caption{Cohen's Kappa for each summary among \textsc{silver} workers (Pairwise).}
\label{fig:S-pair}
\end{figure*}

\vspace{-0.2cm}

\begin{figure*}[!ht]
\begin{center}
\includegraphics[width=1\textwidth]{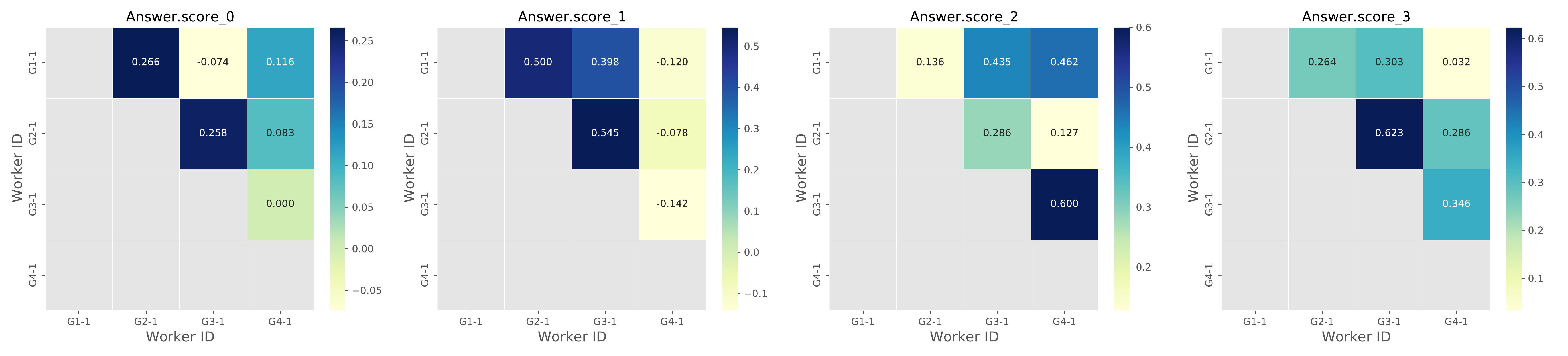}
\end{center}
\vspace{-0.2cm}
\caption{Cohen's Kappa for each summary among \textsc{gold} workers (Pairwise).}
\label{fig:G-pair}
\end{figure*}

\vspace{-0.2cm}

\begin{figure*}[!bht]
\begin{center}
\includegraphics[width=1\textwidth]{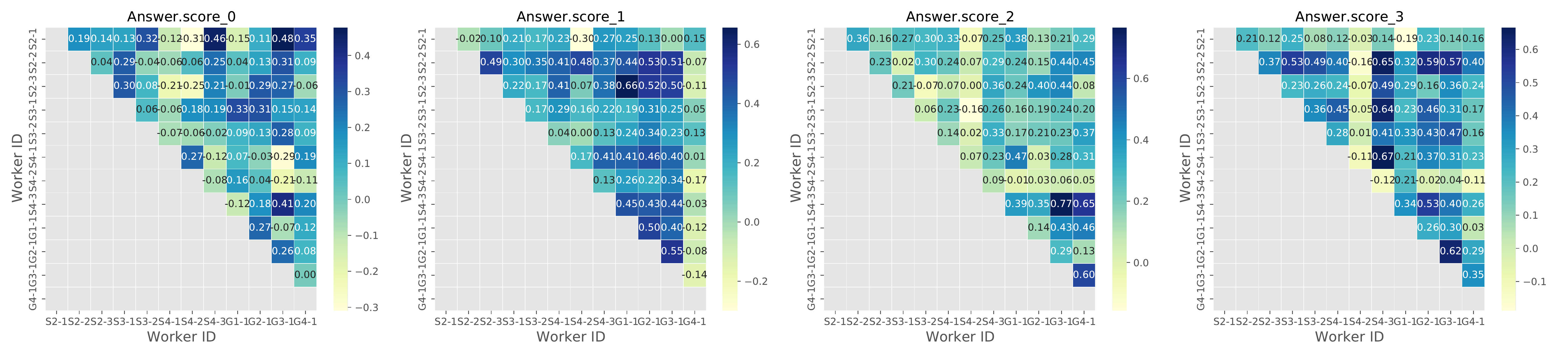}
\end{center}
\vspace{-0.2cm}
\caption{Cohen's Kappa for each summary across \textsc{silver} and \textsc{gold} workers (Pairwise).}
\label{fig:SG-pair}
\end{figure*}

\clearpage
\subsection{Endurance Task Result of Lab Members}
\label{sec:app-enduranceLab}
\begin{table*}[!htb]
  \centering
  \begin{tabular}{ccccc}
    \toprule
    \multicolumn{2}{c}{Worker Combination} & 
    A and B & B and C & C and A \\
    \hline
    \multirow{4}*{Cohen's Kappa (Each Summary)} 
    & Answer.score\_0 & -0.261 & -0.083 & 0.246 \\
    & Answer.score\_1 & 0.285 & 0.13 & 0.285 \\
    & Answer.score\_2 & 0.206 & -0.006 & -0.049 \\
    & Answer.score\_3 & 0.066 & 0.006 & 0.387 \\
    \hline
    \multicolumn{2}{c}{Cohen's Kappa (Concatenation)} 
    & 0.1 & 0.055 & 0.268 \\
    \hline
    \multicolumn{2}{c}{Cohen's Kappa (Omit first 2 HITs)} & 0.2 & 0.091 & 0.196 \\
    \hline
    \multicolumn{2}{c}{Krippendorff's Alpha} & \multicolumn{3}{c}{0.201} \\
    \bottomrule
  \end{tabular}
  \caption{Endurance task result of lab members.}
  \label{tab:lab}
\end{table*}

\subsection{Statistical Test Results of Qualification and Endurance Tasks for Each Round}
\label{sec:app-statTestRound}

\begin{table*}[!htb]
  \label{tab:stat2}
  \centering
  \begin{tabular}{ccccc}
    \toprule
    \multicolumn{2}{c}{Annotation Task} 
    & \makecell{Pass Rate}
    & \makecell{Mean of \\ Pass Rate (Bootstrap)}
    & \makecell{Standard Dev. of \\ Pass Rate (Bootstrap)} \\
    \hline
    \multirow{2}{*}{Round 1}
    & Qua. Task & 0.1 & 0.0997 & 0.0424 \\
    & End. Task & 0.02 & 0.0199 & 0.0198 \\
    \hline
    \multirow{2}{*}{Round 2}
    & Qua. Task & 0.16 & 0.1611 & 0.0521 \\
    & End. Task & 0.08 & 0.0805 & 0.0384 \\
    \hline
    \multirow{2}{*}{Round 3}
    & Qua. Task & 0.1 & 0.1000 & 0.0482 \\
    & End. Task & 0.06 & 0.0599 & 0.0339 \\
    \hline
    \multirow{2}{*}{Round 4}
    & Qua. Task & 0.16 & 0.1595 & 0.0511 \\
    & End. Task & 0.08 & 0.0800 & 0.0380 \\
    \hline
    \multirow{2}{*}{All Rounds}
    & Qua. Task & 0.13 & 0.1302 & 0.0236 \\
    & End. Task & 0.06 & 0.0602 & 0.0168 \\
    \bottomrule
  \end{tabular}
  \caption{Statistical test results of qualification and endurance task.}
\end{table*}

\subsection{Cohen's Kappa (combined scores) for CloudResearch Workers in Reference-based Task}

\begin{figure}[!h]
\begin{center}
\includegraphics[width=0.4\textwidth]{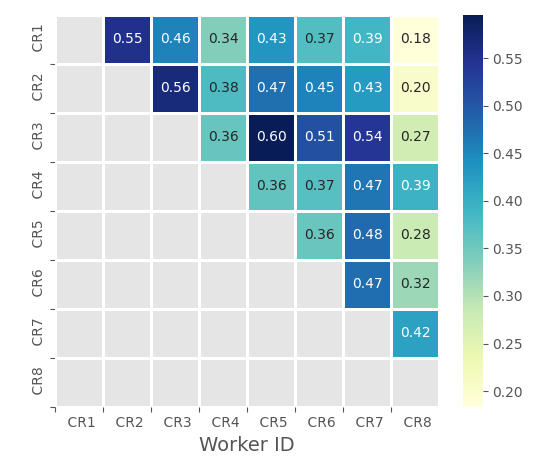}
\end{center}
\caption{Cohen's Kappa (combined scores) among CloudResearch workers.}
\label{fig:ref-cloudresearch}
\end{figure}

\clearpage
\subsection{Spearman's Coefficient for Inter-groups (Pipeline \& MACE) in Reference-based Task}
For the reference-based task, we used 4 methods to calculate Spearman’s coefficient:

\begin{itemize}
  \item \textbf{Method 1}: Given the different numbers of remaining MACE workers for each HIT, we calculate Spearman’s coefficient between our pipeline and MACE workers in each HIT. Then we take the average of these coefficients as the inter-group Spearman’s coefficient shown in Table \ref{tab:MACE-inter} \footnote{We also calculate Spearman’s coefficient within the group of our pipeline and MACE workers separately for comparison, as shown in Table \ref{tab:MACE-inter}.}.
  \vspace*{-3mm}
  \item \textbf{Method 2}: The only difference between this method and Method 1 is that we take the absolute value when calculating Spearman’s coefficient for each HIT.
  \vspace*{-3mm}
  \item \textbf{Method 3}: We take the average of each annotation question in each HIT within the group of our pipeline and MACE workers separately, then concatenate these average scores of all HITs together for each group and calculate Spearman’s coefficient. 
  \vspace*{-3mm}
  \item \textbf{Method 4}: The only difference between this method and Method 3 is that we calculate Spearman’s coefficient for each HIT and then take the average of all coefficients instead of concatenating first and then calculating the coefficient.
\end{itemize}

\begin{table}[bht]
  \centering
  \resizebox{0.5\textwidth}{!}{
  \begin{tabular}{ccccc}
    \toprule
    \multicolumn{2}{c}{Threshold} & 0.5 & 0.6 & 0.7 \\
    \hline
    \multicolumn{2}{c}{\% of workers kept} & 19.2\% & 15.9\% & 7.6\% \\
    \multicolumn{2}{c}{HIT coverage} & 30/30 & 27/30 & 18/30 \\
    \multicolumn{2}{c}{Avg. num. workers per HIT} & 2.4 & 1.9 & 1.2 \\
    \hline
    \multicolumn{2}{c}{\makecell{Krippendorff’s Alpha \\ (all scores)}} & 0.380 & 0.472 & 0.754 \\

    % 1) w/o absolute value
    \hline
    \multirow{3}{*}{Method 1} 
    & \makecell{Spearman's coefficient \\ (MACE workers)} & 0.351 & 0.414 & 0.770 \\
    & \makecell{Spearman's coefficient \\ (pipeline workers)} & 0.558 & 0.565 & 0.577 \\
    & \makecell{Spearman's coefficient \\ (inter-group)} & -0.081 & -0.063 & -0.234 \\
    
    % 2) w/ absolute value
    \hline
    \multirow{3}{*}{Method 2} 
    & \makecell{Spearman's coefficient \\ (MACE workers)} & 0.396 & 0.418 & 0.770 \\
    & \makecell{Spearman's coefficient \\ (pipeline workers)} & 0.575 & 0.580 & 0.591 \\
    & \makecell{Spearman's coefficient \\ (inter-group)} & 0.307 & 0.299 & 0.308 \\
    
    % 3) avg. annotations + concatenate + calculate Spearman's
    \hline
    \multirow{1}{*}{Method 3} 
    & \makecell{Spearman's coefficient \\ (inter-group)} & -0.107 & -0.067 & -0.355 \\
    
    % 4) avg. annotations + calculate Spearman's + avg. Spearman's
    \hline
    \multirow{1}{*}{Method 4} 
    & \makecell{Spearman's coefficient \\ (inter-group)} & -0.102 & -0.113 & -0.194 \\
    
    % \makecell{Num. of remained HITs \\ for coefficient calculation} & 24 & 17 & 3 \\
    % \makecell{Avg. Num. of workers per HIT \\ for coefficient calculation} & 2.8 & 2.4 & 2.0 \\
    \bottomrule
  \end{tabular}}
  \caption{Methods for calculation of Spearman's coefficient within and across groups of pipeline and MACE workers in reference-based task.}
  \label{tab:MACE-inter}
\end{table}

\clearpage
\subsection{Qualitative Analysis of Correctness Across Annotation Sources in Reference-based Task}
\label{sec:50-samples}
For the reference-based task, we first randomly select 50 HITs out of 30 HITs (HIT index ranges from 0 to 29), and then 1 annotation question out of 8 questions (annotation index ranges from 0 to 7) for each of these HITs selected in the above step. 

For each randomly selected annotation question, we calculate the median within the groups of our pipeline, MACE, and CloudResearch workers separately, as well as the scores generated by GPT models (GPT-3.5 (\say{text-davinci-003}), ChatGPT, and GPT-4\footnote{For the ChatGPT score, we ran 5 times with default parameters (temperature=1, top\_p=1) to take the median, but set the temperature as 0 with a single run for GPT-3.5 and GPT-4 scores.}). The expert judgment (aggregated by the median) and details for 50 randomly selected annotation questions can be found in Table \ref{tab:random-check-sample-p1} and Table \ref{tab:random-check-sample-p2}.

\begin{figure*}[!thb]
\begin{center}
\includegraphics[width=0.48\textwidth]{images/heat_spearsman_coeff.png}~~~
\includegraphics[width=0.48\textwidth]{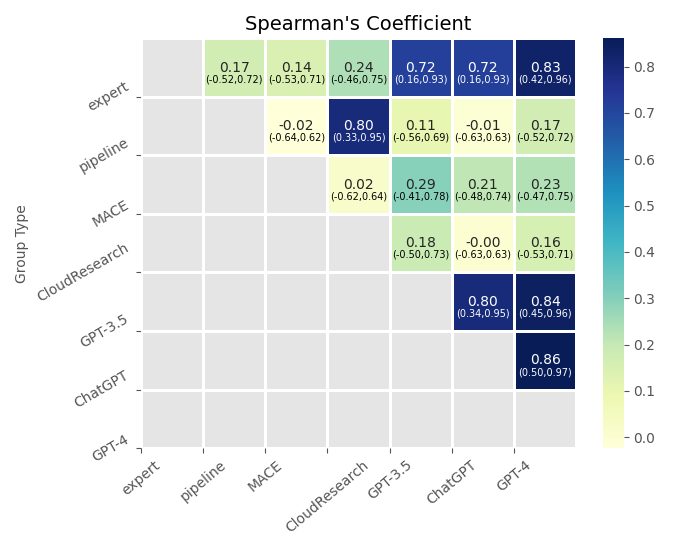}
\end{center}
\caption{Spearman’s coefficient for scores of 50 random samples aggregated by \textbf{median} among groups before (left) and after (right) the removal of controversial HITs (95\% confidence interval is shown below the coefficient).}
\label{fig:heat-spearman-median-2fig}
\end{figure*}

\begin{figure*}[!thb]
\begin{center}
\includegraphics[width=0.48\textwidth]{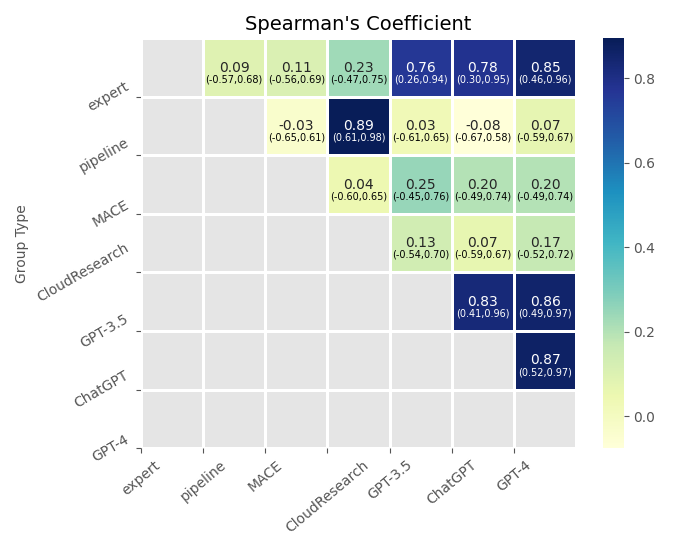}~~~
\includegraphics[width=0.48\textwidth]{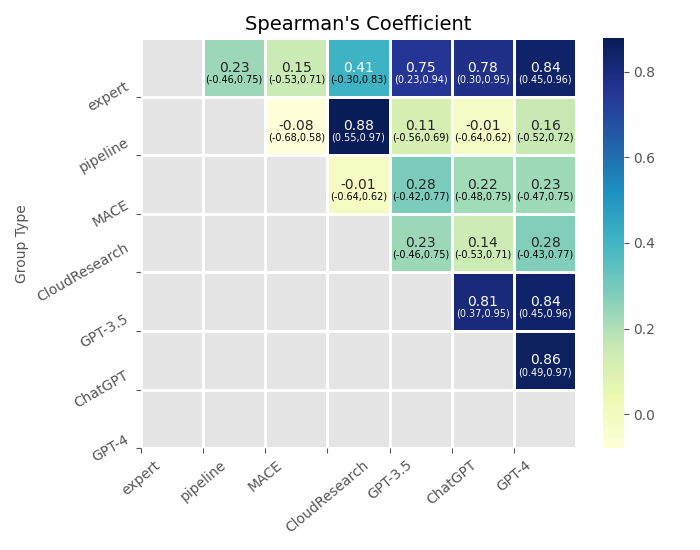}
\end{center}
\caption{Spearman’s coefficient for scores of 50 random samples aggregated by \textbf{mean} among groups before (left) and after (right) the removal of controversial HITs (95\% confidence interval is shown below the coefficient).}
\label{fig:heat-spearman-mean-2fig}
\end{figure*}

Figure \ref{fig:heat-spearman-median-2fig} shows Spearman’s coefficient among different groups aggregated by the median before (left) and after (right) the removal of controversial HITs (HIT with index 15, 16, and 28). 
We also perform a similar analysis aggregated by the mean shown in Figure \ref{fig:heat-spearman-mean-2fig}. 

\begin{table}[h!]
\begin{threeparttable}
  \centering
  \resizebox{1\textwidth}{!}{
  \begin{tabular}{c|cc|c|ccc|ccc|c}
    \toprule
    \multirow{2}{*}{\makecell{Sample\\Index}} & 
    \multicolumn{2}{c|}{Two Types of Summaries} & 
    \multirow{2}{*}{\makecell{Inclusion\\Direction}} & 
    \multicolumn{3}{c|}{Human Annotators (Median)} & 
    \multicolumn{3}{c|}{GPT series scores} & 
    \multirow{2}*{\makecell{Expert\\Judgment}} \\
    & & & & Pipeline & MACE & CloudResearch & GPT-3.5 & ChatGPT & GPT-4 \\
    \hline
    \multirow{2}{*}{1} & 
    Reference & 
    \makecell[l]{The government has given regulators more time to investigate the proposed takeover of \\broadcaster Sky by 21st Century Fox.} &
    \multirow{2}{*}{can2ref} &
    \multirow{2}{*}{5.0} &
    \multirow{2}{*}{4.0} &
    \multirow{2}{*}{4.0} & 
    \multirow{2}{*}{4.0} & 
    \multirow{2}{*}{5.0} & 
    \multirow{2}{*}{5.0} &
    \multirow{2}{*}{5.0} \\
    & Candidate & 
    \makecell[l]{The government has extended the deadline for an inquiry into the takeover of Sky by 21st \\Century Fox.} & & & & & & & & \\
    \hline
    \multirow{2}{*}{2} & 
    Reference & 
    \makecell[l]{A Chinese woman has been found guilty of trespassing at President Donald Trump's \\Mar-a-Lago club in Florida and of lying to a federal agent.} & 
    \multirow{2}{*}{can2ref} &
    \multirow{2}{*}{3.0} & 
    \multirow{2}{*}{5.0} & 
    \multirow{2}{*}{3.5} & 
    \multirow{2}{*}{4} & 
    \multirow{2}{*}{5} & 
    \multirow{2}{*}{4.5} &  
    \multirow{2}{*}{4.0} \\
    & Candidate & 
    \makecell[l]{A Chinese woman who sparked alarm when she walked into US President Donald Trump's \\Mar-a-Lago resort has been found guilty of trespassing.} & & & & & & & & \\
    \hline
    \multirow{2}{*}{3} & 
    Reference & 
    \makecell[l]{A unique garden is helping Canadians to break a taboo that exists in many societies. It is \\allowing parents to talk openly about miscarriage.} & 
    \multirow{2}{*}{ref2can} &
    \multirow{2}{*}{4.0} & 
    \multirow{2}{*}{4.0} & 
    \multirow{2}{*}{4.0} & 
    \multirow{2}{*}{4.0} & 
    \multirow{2}{*}{5.0} & 
    \multirow{2}{*}{4.0} & 
    \multirow{2}{*}{3.0} \\
    & Candidate & 
    \makecell[l]{A Canadian cemetery has created a garden dedicated to the memory of babies lost during \\pregnancy. It's a place that's especially for those who have had multiple miscarriages.} & & & & & & & & \\
    \hline
    \multirow{2}{*}{4} & 
    Reference & 
    \makecell[l]{Gadgets that track your steps, sleeping and heart rate could help us live longer and cut \\national healthcare costs by billions - or so we are told.} & 
    \multirow{2}{*}{can2ref} & 
    \multirow{2}{*}{3.0} & 
    \multirow{2}{*}{4.0} & 
    \multirow{2}{*}{2.5} & 
    \multirow{2}{*}{1.0} & 
    \multirow{2}{*}{1.0} & 
    \multirow{2}{*}{1.0} & 
    \multirow{2}{*}{1.0} \\
    & Candidate & 
    \makecell[l]{It is a huge amount of us have a smartphone, a smartphone and a gadget that feeds data \\from a smartphone.} & & & & & & & & \\
    \hline
    \multirow{2}{*}{5} & 
    Reference & 
    \makecell[l]{A unique garden is helping Canadians to break a taboo that exists in many societies. It is \\allowing parents to talk openly about miscarriage.} & 
    \multirow{2}{*}{can2ref} & 
    \multirow{2}{*}{2.0} & 
    \multirow{2}{*}{4.0} & 
    \multirow{2}{*}{2.0} & 
    \multirow{2}{*}{4.0} & 
    \multirow{2}{*}{5.0} & 
    \multirow{2}{*}{4.0} & 
    \multirow{2}{*}{3.0} \\
    & Candidate & 
    \makecell[l]{A Canadian garden dedicated to the memory of children lost during pregnancy is helping \\to heal the pain of grief.} & & & & & & & & \\
    \hline
    \multirow{2}{*}{6} & 
    Reference & 
    \makecell[l]{The 2017 Oscar nominations are out, with La La Land the frontrunner. Here's a round-up \\of the surprises and talking points from this year's list.} &
    \multirow{2}{*}{can2ref} & 
    \multirow{2}{*}{4.0} & 
    \multirow{2}{*}{3.0} & 
    \multirow{2}{*}{3.5} & 
    \multirow{2}{*}{3.0} & 
    \multirow{2}{*}{4.0} & 
    \multirow{2}{*}{4.0} & 
    \multirow{2}{*}{4.0} \\
    & Candidate & 
    \makecell[l]{The full list of Oscar nominations has been announced. Here are 10 talking points from \\the shortlists.} & & & & & & & & \\
    \hline
    \multirow{2}{*}{7} & 
    Reference &
    \makecell[l]{Welsh victims of the contaminated blood scandal have said it is not fair they get less \\financial help than people affected in England and Scotland.} & 
    \multirow{2}{*}{ref2can} & 
    \multirow{2}{*}{2.0} & 
    \multirow{2}{*}{4.0} & 
    \multirow{2}{*}{1.5} & 
    \multirow{2}{*}{4.0} & 
    \multirow{2}{*}{4.0} & 
    \multirow{2}{*}{4.0} & 
    \multirow{2}{*}{2.0} \\
    & Candidate & 
    \makecell[l]{A man who contracted hepatitis C from the contaminated blood scandal has said Welsh \\support payments are not fair.} & & & & & & & & \\
    \hline
    \multirow{2}{*}{8} & 
    Reference &
    \makecell[l]{An anonymous letter sent to a council outlining an alleged plan to oust head teachers is \\"defamatory", the leader of Birmingham City Council has said.} & 
    \multirow{2}{*}{can2ref} & 
    \multirow{2}{*}{4.0} & 
    \multirow{2}{*}{4.0} & 
    \multirow{2}{*}{4.0} & 
    \multirow{2}{*}{3.0} & 
    \multirow{2}{*}{1.0} & 
    \multirow{2}{*}{3.0} & 
    \multirow{2}{*}{2.0} \\
    & Candidate &
    \makecell[l]{A letter written by a council officer calling for schools to be taken over by a council has \\been defamatory.} & & & & & & & & \\
    \hline
    \multirow{2}{*}{9} & 
    Reference &
    \makecell[l]{Graduates from ethnic minorities in Britain are less likely to be in work than their white \\peers, a study says.} & 
    \multirow{2}{*}{ref2can} & 
    \multirow{2}{*}{4.0} & 
    \multirow{2}{*}{4.0} & 
    \multirow{2}{*}{3.5} & 
    \multirow{2}{*}{2.0} & 
    \multirow{2}{*}{2.0} & 
    \multirow{2}{*}{1.0} & 
    \multirow{2}{*}{2.0} \\
    & Candidate &
    \makecell[l]{The number of ethnic minority graduates in the UK has fallen by almost 5\% in the last \\year, according to a think tank.} & & & & & & & & \\
    \hline
    \multirow{2}{*}{10} & 
    Reference &
    \makecell[l]{Two endangered red panda cubs have been born at a wildlife park on the Isle of Man.} & 
    \multirow{2}{*}{ref2can} & 
    \multirow{2}{*}{2.0} & 
    \multirow{2}{*}{4.0} & 
    \multirow{2}{*}{4.0} & 
    \multirow{2}{*}{5.0} & 
    \multirow{2}{*}{5.0} & 
    \multirow{2}{*}{5.0} & 
    \multirow{2}{*}{5.0} \\
    & Candidate &
    \makecell[l]{Two endangered red panda cubs have been born at a wildlife park in the Isle of Man.} & & & & & & & & \\
    \hline
    \multirow{2}{*}{11} & 
    Reference &
    \makecell[l]{Two endangered red panda cubs have been born at a wildlife park on the Isle of Man.} & 
    \multirow{2}{*}{ref2can} & 
    \multirow{2}{*}{5.0} & 
    \multirow{2}{*}{3.0} & 
    \multirow{2}{*}{5.0} & 
    \multirow{2}{*}{4.0} & 
    \multirow{2}{*}{4.0} & 
    \multirow{2}{*}{5.0} & 
    \multirow{2}{*}{5.0} \\
    & Candidate &
    \makecell[l]{Two endangered red panda cubs have been born at a wildlife park on the Isle of Man, \\a year after a giant themed elephant calf escaped from his enclosure.} & & & & & & & & \\
    \hline
    \multirow{2}{*}{12} & 
    Reference &
    \makecell[l]{Welsh Water has announced pre-tax profits of £7m for the last financial year.} & 
    \multirow{2}{*}{can2ref} & 
    \multirow{2}{*}{5.0} & 
    \multirow{2}{*}{4.0} & 
    \multirow{2}{*}{5.0} & 
    \multirow{2}{*}{5.0} & 
    \multirow{2}{*}{5.0} & 
    \multirow{2}{*}{5.0} & 
    \multirow{2}{*}{4.0} \\
    & Candidate &
    \makecell[l]{Welsh Water has announced pre-tax profits of £7m for the year to April.} & & & & & & & & \\
    \hline
    \multirow{2}{*}{13} & 
    Reference &
    \makecell[l]{A "poo-powered" VW Beetle has taken to the streets of Bristol in an attempt to \\encourage sustainable motoring.} & 
    \multirow{2}{*}{ref2can} & 
    \multirow{2}{*}{4.0} & 
    \multirow{2}{*}{4.0} & 
    \multirow{2}{*}{2.5} & 
    \multirow{2}{*}{4.0} & 
    \multirow{2}{*}{4.0} & 
    \multirow{2}{*}{4.0} & 
    \multirow{2}{*}{3.0} \\
    & Candidate &
    \makecell[l]{A car powered by biogas has been seen on the streets of Bristol.} & & & & & & & & \\
    \hline
    \multirow{2}{*}{14} & 
    Reference &
    \makecell[l]{An anonymous letter sent to a council outlining an alleged plan to oust head teachers \\is "defamatory", the leader of Birmingham City Council has said.} & 
    \multirow{2}{*}{can2ref} & 
    \multirow{2}{*}{5.0} & 
    \multirow{2}{*}{3.0} & 
    \multirow{2}{*}{4.5} & 
    \multirow{2}{*}{4.0} & 
    \multirow{2}{*}{5.0} & 
    \multirow{2}{*}{4.0} & 
    \multirow{2}{*}{3.0} \\
    & Candidate &
    \makecell[l]{A letter sent to Birmingham City Council by a whistle-blower has been described as \\"defamatory" by the city council's chief inspector of schools.} & & & & & & & & \\
    \hline
    \multirow{2}{*}{15} & 
    Reference &
    \makecell[l]{In our media-saturated age, it's rare to have a chief executive who doesn't speak to the \\press or, indeed, very often publicly.} & 
    \multirow{2}{*}{can2ref} & 
    \multirow{2}{*}{5.0} & 
    \multirow{2}{*}{4.0} & 
    \multirow{2}{*}{5.0} & 
    \multirow{2}{*}{1.0} & 
    \multirow{2}{*}{1.0} & 
    \multirow{2}{*}{1.0} & 
    \multirow{2}{*}{1.0} \\
    & Candidate &
    \makecell[l]{Chinese entrepreneurs are a familiar sight.} & & & & & & & & \\
    \hline
    \multirow{2}{*}{16} & 
    Reference &
    \makecell[l]{Parliament has been dissolved and the official election campaign has begun. BBC Reality \\Check listened in to Prime Minister Boris Johnson's campaign speeches in Downing Street \\and in Birmingham to check the facts and figures.} & 
    \multirow{2}{*}{ref2can} & 
    \multirow{2}{*}{5.0} & 
    \multirow{2}{*}{4.0} & 
    \multirow{2}{*}{5.0} & 
    \multirow{2}{*}{3.0} & 
    \multirow{2}{*}{3.0} & 
    \multirow{2}{*}{3.0} & 
    \multirow{2}{*}{1.0} \\
    & Candidate &
    \makecell[l]{Boris Johnson made a series of claims about his government's plans for the next few years. \\Here are six of the key pledges he made.} & & & & & & & & \\
    \hline
    \multirow{2}{*}{17} & 
    Reference &
    \makecell[l]{Naturalist Sir David Attenborough and the Queen are the greatest living British man and \\woman, according to readers of Best of British magazine.} & 
    \multirow{2}{*}{can2ref} & 
    \multirow{2}{*}{3.0} & 
    \multirow{2}{*}{4.0} & 
    \multirow{2}{*}{3.5} & 
    \multirow{2}{*}{4.0} & 
    \multirow{2}{*}{5.0} & 
    \multirow{2}{*}{4.0} & 
    \multirow{2}{*}{4.0} \\
    & Candidate &
    \makecell[l]{David Attenborough has been voted the best of British by the magazine.} & & & & & & & & \\
    \hline
    \multirow{2}{*}{18} & 
    Reference &
    \makecell[l]{An Edinburgh adventurer has become the youngest woman to ski solo to the South Pole.} & 
    \multirow{2}{*}{can2ref} & 
    \multirow{2}{*}{4.0} & 
    \multirow{2}{*}{4.0} & 
    \multirow{2}{*}{4.0} & 
    \multirow{2}{*}{5.0} & 
    \multirow{2}{*}{5.0} & 
    \multirow{2}{*}{4.5} & 
    \multirow{2}{*}{4.0} \\
    & Candidate &
    \makecell[l]{A woman from Edinburgh has become the youngest person to reach the South Pole solo.} & & & & & & & & \\
    \hline
    \multirow{2}{*}{19} & 
    Reference &
    \makecell[l]{Resurfacing work on a newly-repaired canal towpath that washed away after vandals left \\a lock gate open has begun.} & 
    \multirow{2}{*}{can2ref} & 
    \multirow{2}{*}{4.0} & 
    \multirow{2}{*}{3.0} & 
    \multirow{2}{*}{3.5} & 
    \multirow{2}{*}{4.0} & 
    \multirow{2}{*}{4.0} & 
    \multirow{2}{*}{4.0} & 
    \multirow{2}{*}{5.0} \\
    & Candidate &
    \makecell[l]{Work has begun to resurface a canal towpath which was damaged by flooding.} & & & & & & & & \\
    \hline
    \multirow{2}{*}{20} & 
    Reference &
    \makecell[l]{The Brexit vote is already having a negative impact on business, a survey of bosses from \\some of the UK's biggest companies has suggested.} & 
    \multirow{2}{*}{ref2can} & 
    \multirow{2}{*}{4.0} & 
    \multirow{2}{*}{5.0} & 
    \multirow{2}{*}{4.0} & 
    \multirow{2}{*}{4.0} & 
    \multirow{2}{*}{5.0} & 
    \multirow{2}{*}{4.0} & 
    \multirow{2}{*}{5.0} \\
    & Candidate &
    \makecell[l]{The majority of business leaders believe the Brexit vote has already had a negative impact \\on their company, a survey suggests.} & & & & & & & & \\
    \hline
    \multirow{2}{*}{21} & 
    Reference &
    \makecell[l]{A campaign has begun to stop the spread of norovirus in Cornwall.} & 
    \multirow{2}{*}{can2ref} & 
    \multirow{2}{*}{5.0} & 
    \multirow{2}{*}{5.0} & 
    \multirow{2}{*}{3.5} & 
    \multirow{2}{*}{4.0} & 
    \multirow{2}{*}{5.0} & 
    \multirow{2}{*}{5.0} & 
    \multirow{2}{*}{5.0} \\
    & Candidate &
    \makecell[l]{A campaign has been launched to prevent the spread of norovirus in Cornwall.} & & & & & & & & \\
    \hline
    \multirow{2}{*}{22} & 
    Reference &
    \makecell[l]{Welsh victims of the contaminated blood scandal have said it is not fair they get less \\financial help than people affected in England and Scotland.} & 
    \multirow{2}{*}{can2ref} & 
    \multirow{2}{*}{3.0} & 
    \multirow{2}{*}{3.5} & 
    \multirow{2}{*}{2.5} & 
    \multirow{2}{*}{3.0} & 
    \multirow{2}{*}{2.0} & 
    \multirow{2}{*}{2.0} & 
    \multirow{2}{*}{3.0} \\
    & Candidate &
    \makecell[l]{The Welsh Government has said it is not fair to pay for patients who have contaminated \\blood in the 1970s and 1980s.} & & & & & & & & \\
    \hline
    \multirow{2}{*}{23} & 
    Reference &
    \makecell[l]{People on Jersey's Ecrehous islands are concerned travellers are arriving from France by \\boat and not being tested for coronavirus.} & 
    \multirow{2}{*}{ref2can} & 
    \multirow{2}{*}{1.0} & 
    \multirow{2}{*}{4.0} & 
    \multirow{2}{*}{1.0} & 
    \multirow{2}{*}{3.0} & 
    \multirow{2}{*}{3.0} & 
    \multirow{2}{*}{3.0} & 
    \multirow{2}{*}{2.0} \\
    & Candidate &
    \makecell[l]{People living on Jersey's Ecrehous islands have said they are worried about the number \\of people arriving ashore.} & & & & & & & & \\
    \hline
    \multirow{2}{*}{24} & 
    Reference &
    \makecell[l]{The government has given regulators more time to investigate the proposed takeover of \\broadcaster Sky by 21st Century Fox.} & 
    \multirow{2}{*}{can2ref} & 
    \multirow{2}{*}{2.5} & 
    \multirow{2}{*}{3.0} & 
    \multirow{2}{*}{3.0} & 
    \multirow{2}{*}{4.0} & 
    \multirow{2}{*}{4.0} & 
    \multirow{2}{*}{4.0} & 
    \multirow{2}{*}{3.0} \\
    & Candidate &
    \makecell[l]{The government has extended its takeover inquiry into Sky's takeover deal with regulator \\Ofcom.} & & & & & & & & \\
    \hline
    \multirow{2}{*}{25} & 
    Reference &
    \makecell[l]{Graduates from ethnic minorities in Britain are less likely to be in work than their white \\peers, a study says.} & 
    \multirow{2}{*}{can2ref} & 
    \multirow{2}{*}{3.0} & 
    \multirow{2}{*}{3.5} & 
    \multirow{2}{*}{3.0} & 
    \multirow{2}{*}{2.0} & 
    \multirow{2}{*}{2.0} & 
    \multirow{2}{*}{1.0} & 
    \multirow{2}{*}{2.0} \\
    & Candidate &
    \makecell[l]{The number of ethnic minority graduates in the UK has fallen by almost 5\% in the last \\year, according to a think tank.} & & & & & & & & \\
\bottomrule
  \end{tabular}}
  \tiny  
\end{threeparttable}
  \caption{Qualitative analysis of correctness with 50 random samples (Part 1).}
  \label{tab:random-check-sample-p1}
\end{table}

\begin{table}[h!]
\begin{threeparttable}
  \centering
  \resizebox{1\textwidth}{!}{
  \begin{tabular}{c|cc|c|ccc|ccc|c}
    \toprule
    \multirow{2}{*}{\makecell{Sample\\Index}} & 
    \multicolumn{2}{c|}{Two Types of Summaries} & 
    \multirow{2}{*}{\makecell{Inclusion\\Direction}} & 
    \multicolumn{3}{c|}{Human Annotators (Median)} & 
    \multicolumn{3}{c|}{GPT series scores} & 
    \multirow{2}*{\makecell{Expert\\Judgment}} \\
    & & & & Pipeline & MACE & CloudResearch & GPT-3.5 & ChatGPT & GPT-4 \\
    \hline
    \multirow{2}{*}{26} & 
    Reference &
    \makecell[l]{Joan Miro's 1927 work Peinture (Etoile Bleue) has sold for more than £23.5 million in \\London, setting a new auction record for the Spanish painter.} & 
    \multirow{2}{*}{can2ref} & 
    \multirow{2}{*}{4.0} & 
    \multirow{2}{*}{5.0} & 
    \multirow{2}{*}{4.0} & 
    \multirow{2}{*}{3.0} & 
    \multirow{2}{*}{1.0} & 
    \multirow{2}{*}{2.0} & 
    \multirow{2}{*}{3.0} \\
    & Candidate &
    \makecell[l]{Joan Miro's painting, which inspired the famous Joan Miro, has smashed its auction \\record for £15m.} & & & & & & & & \\
    \hline
    \multirow{2}{*}{27} & 
    Reference &
    \makecell[l]{One of Oxford's main routes remains closed because of flooding for the second time in \\a month.} & 
    \multirow{2}{*}{ref2can} & 
    \multirow{2}{*}{2.5} & 
    \multirow{2}{*}{5.0} & 
    \multirow{2}{*}{2.0} & 
    \multirow{2}{*}{5.0} & 
    \multirow{2}{*}{5.0} & 
    \multirow{2}{*}{5.0} & 
    \multirow{2}{*}{5.0} \\
    & Candidate &
    \makecell[l]{A major route through Oxford has been closed for the second time in a month due to flooding.} & & & & & & & & \\
    \hline
    \multirow{2}{*}{28} & 
    Reference &
    \makecell[l]{Holidaymakers say they have been left thousands of pounds out of pocket after a letting \\company ceased trading without notice.} & 
    \multirow{2}{*}{ref2can} & 
    \multirow{2}{*}{5.0} & 
    \multirow{2}{*}{4.0} & 
    \multirow{2}{*}{5.0} & 
    \multirow{2}{*}{4.0} & 
    \multirow{2}{*}{4.0} & 
    \multirow{2}{*}{4.0} & 
    \multirow{2}{*}{2.0} \\
    & Candidate &
    \makecell[l]{Brighton Holiday Homes has gone bust with bookings cancelled after a third of its customers \\claimed their money was lost.} & & & & & & & & \\
    \hline
    \multirow{2}{*}{29} & Reference & \makecell[l]{A £4.4m revamped Denbighshire leisure centre will open on Saturday.}	&  \multirow{2}{*}{cand2ref}   &  \multirow{2}{*}{4.0}   &  \multirow{2}{*}{5.0}   &  \multirow{2}{*}{3.0}   &  \multirow{2}{*}{5.0}   &  \multirow{2}{*}{4.0}   &  \multirow{2}{*}{4.0}   &  \multirow{2}{*}{4.0}	\\ & Candidate & \makecell[l]{A Denbighshire leisure centre is reopening on Thursday after a £4.4m revamp.} & & & & & & & & \\ \hline
    \multirow{2}{*}{30} & Reference & \makecell[l]{Gadgets that track your steps, sleeping and heart rate could help us live longer and cut national \\healthcare costs by billions - or so we are told.}	&  \multirow{2}{*}{ref2cand}   &  \multirow{2}{*}{1.0}   &  \multirow{2}{*}{3.5}   &  \multirow{2}{*}{3.0}   &  \multirow{2}{*}{4.0}   &  \multirow{2}{*}{1.0}   &  \multirow{2}{*}{1.0}   &  \multirow{2}{*}{1.0}	\\ & Candidate & \makecell[l]{Every step we take is going to be tracked by a device that cannot simply put our fingers on our wrists.} & & & & & & & & \\ \hline
    \multirow{2}{*}{31} & Reference & \makecell[l]{Gadgets that track your steps, sleeping and heart rate could help us live longer and cut national \\healthcare costs by billions - or so we are told.}	&  \multirow{2}{*}{cand2ref}   &  \multirow{2}{*}{2.0}   &  \multirow{2}{*}{3.0}   &  \multirow{2}{*}{4.0}   &  \multirow{2}{*}{4.0}   &  \multirow{2}{*}{1.0}   &  \multirow{2}{*}{1.0}   &  \multirow{2}{*}{1.0}	\\ & Candidate & \makecell[l]{Every step we take is going to be tracked by a device that cannot simply put our fingers on our wrists.} & & & & & & & & \\ \hline
    \multirow{2}{*}{32} & Reference & \makecell[l]{Joan Miro's 1927 work Peinture (Etoile Bleue) has sold for more than £23.5 million in London, \\setting a new auction record for the Spanish painter.}	&  \multirow{2}{*}{ref2cand}   &  \multirow{2}{*}{2.0}   &  \multirow{2}{*}{5.0}   &  \multirow{2}{*}{4.0}   &  \multirow{2}{*}{4.5}   &  \multirow{2}{*}{4.0}   &  \multirow{2}{*}{4.0}   &  \multirow{2}{*}{4.0}	\\ & Candidate & \makecell[l]{A painting by Joan Miro has sold for £18.8m at auction, breaking the previous record for a \\work by the artist.} & & & & & & & & \\ \hline
    \multirow{2}{*}{33} & Reference & \makecell[l]{A unique garden is helping Canadians to break a taboo that exists in many societies. It is \\allowing parents to talk openly about miscarriage.}	&  \multirow{2}{*}{cand2ref}   &  \multirow{2}{*}{3.0}   &  \multirow{2}{*}{3.0}   &  \multirow{2}{*}{4.0}   &  \multirow{2}{*}{3.0}   &  \multirow{2}{*}{4.0}   &  \multirow{2}{*}{5.0}   &  \multirow{2}{*}{5.0}	\\ & Candidate & \makecell[l]{A Canadian memorial garden is helping parents come to terms with the pain of losing \\a child during pregnancy.} & & & & & & & & \\ \hline
    \multirow{2}{*}{34} & Reference & \makecell[l]{Holidaymakers say they have been left thousands of pounds out of pocket after a letting \\company ceased trading without notice.}	&  \multirow{2}{*}{cand2ref}   &  \multirow{2}{*}{3.0}   &  \multirow{2}{*}{4.0}   &  \multirow{2}{*}{4.0}   &  \multirow{2}{*}{4.0}   &  \multirow{2}{*}{4.0}   &  \multirow{2}{*}{3.0}   &  \multirow{2}{*}{4.0}	\\ & Candidate & \makecell[l]{A holiday home firm has gone bust after customers were told they had been left "heartbroken" \\after bookings were cancelled.} & & & & & & & & \\ \hline
    \multirow{2}{*}{35} & Reference & \makecell[l]{A woman rescued after falling from a North Sea ferry has told how she thought she was going to die.}	&  \multirow{2}{*}{ref2cand}   &  \multirow{2}{*}{4.0}   &  \multirow{2}{*}{4.5}   &  \multirow{2}{*}{5.0}   &  \multirow{2}{*}{1.5}   &  \multirow{2}{*}{5.0}   &  \multirow{2}{*}{5.0}   &  \multirow{2}{*}{5.0}	\\ & Candidate & \makecell[l]{A woman who fell from a ferry into the North Sea has described how she thought she was going to die.} & & & & & & & & \\ \hline
    \multirow{2}{*}{36} & Reference & \makecell[l]{The Brexit vote is already having a negative impact on business, a survey of bosses from \\some of the UK's biggest companies has suggested.}	&  \multirow{2}{*}{ref2cand}   &  \multirow{2}{*}{4.0}   &  \multirow{2}{*}{3.0}   &  \multirow{2}{*}{3.0}   &  \multirow{2}{*}{2.0}   &  \multirow{2}{*}{4.0}   &  \multirow{2}{*}{5.0}   &  \multirow{2}{*}{5.0}	\\ & Candidate & \makecell[l]{The UK's vote to leave the European Union is already having a negative impact on businesses, \\a survey suggests.} & & & & & & & & \\ \hline
    \multirow{2}{*}{37} & Reference & \makecell[l]{Welsh Water has announced pre-tax profits of £7m for the last financial year.}	&  \multirow{2}{*}{ref2cand}   &  \multirow{2}{*}{4.0}   &  \multirow{2}{*}{4.5}   &  \multirow{2}{*}{4.0}   &  \multirow{2}{*}{4.0}   &  \multirow{2}{*}{5.0}   &  \multirow{2}{*}{5.0}   &  \multirow{2}{*}{5.0}	\\ & Candidate & \makecell[l]{Welsh Water has announced pre-tax profits of £7m for the year to April.} & & & & & & & & \\ \hline
    \multirow{2}{*}{38} & Reference & \makecell[l]{One of Oxford's main routes remains closed because of flooding for the second time in a month.}	&  \multirow{2}{*}{ref2cand}   &  \multirow{2}{*}{5.0}   &  \multirow{2}{*}{3.0}   &  \multirow{2}{*}{5.0}   &  \multirow{2}{*}{3.5}   &  \multirow{2}{*}{5.0}   &  \multirow{2}{*}{5.0}   &  \multirow{2}{*}{5.0}	\\ & Candidate & \makecell[l]{A major route through Oxford has been closed for the second time in a month because of flooding.} & & & & & & & & \\ \hline
    \multirow{2}{*}{39} & Reference & \makecell[l]{A 10-year-old boy died after he hit his head on a wall while playing football at school, an inquest heard.}	&  \multirow{2}{*}{ref2cand}   &  \multirow{2}{*}{4.0}   &  \multirow{2}{*}{4.0}   &  \multirow{2}{*}{3.0}   &  \multirow{2}{*}{3.5}   &  \multirow{2}{*}{4.0}   &  \multirow{2}{*}{5.0}   &  \multirow{2}{*}{5.0}	\\ & Candidate & \makecell[l]{A 10-year-old boy who hit his head while playing football at school died from traumatic brain injury, \\an inquest heard.} & & & & & & & & \\ \hline
    \multirow{2}{*}{40} & Reference & \makecell[l]{A video artist who uses YouTube clips, a print-maker and an artist who pairs spoken word with \\photography are among this year's Turner Prize nominees.}	&  \multirow{2}{*}{ref2cand}   &  \multirow{2}{*}{3.0}   &  \multirow{2}{*}{4.0}   &  \multirow{2}{*}{3.5}   &  \multirow{2}{*}{3.5}   &  \multirow{2}{*}{4.0}   &  \multirow{2}{*}{4.0}   &  \multirow{2}{*}{4.0}	\\ & Candidate & \makecell[l]{A YouTube artist who splices together clips of horror films and a print-maker who works with \\women's groups are among the nominees for this year's Turner Prize.} & & & & & & & & \\ \hline
    \multirow{2}{*}{41} & Reference & \makecell[l]{Parliament has been dissolved and the official election campaign has begun. BBC Reality Check \\listened in to Prime Minister Boris Johnson's campaign speeches in Downing Street and in \\Birmingham to check the facts and figures.}	&  \multirow{2}{*}{ref2cand}   &  \multirow{2}{*}{1.0}   &  \multirow{2}{*}{1.0}   &  \multirow{2}{*}{3.0}   &  \multirow{2}{*}{1.0}   &  \multirow{2}{*}{3.0}   &  \multirow{2}{*}{4.0}   &  \multirow{2}{*}{2.0}	\\ & Candidate & \makecell[l]{Boris Johnson has been making his pitch to Conservative voters in the final week of the election \\campaign. What did he get right and wrong?} & & & & & & & & \\ \hline
    \multirow{2}{*}{42} & Reference & \makecell[l]{Film director Roman Polanski has been released after being questioned by prosecutors in Poland \\over sex offences in the US.}	&  \multirow{2}{*}{cand2ref}   &  \multirow{2}{*}{3.0}   &  \multirow{2}{*}{4.0}   &  \multirow{2}{*}{3.5}   &  \multirow{2}{*}{4.0}   &  \multirow{2}{*}{4.0}   &  \multirow{2}{*}{4.0}   &  \multirow{2}{*}{4.0}	\\ & Candidate & \makecell[l]{Polish film director Roman Polanski has been freed after prosecutors said they had not made an \\extradition bid for him.} & & & & & & & & \\ \hline
    \multirow{2}{*}{43} & Reference & \makecell[l]{A video artist who uses YouTube clips, a print-maker and an artist who pairs spoken word with \\photography are among this year's Turner Prize nominees.}	&  \multirow{2}{*}{cand2ref}   &  \multirow{2}{*}{3.0}   &  \multirow{2}{*}{3.5}   &  \multirow{2}{*}{4.0}   &  \multirow{2}{*}{4.0}   &  \multirow{2}{*}{3.0}   &  \multirow{2}{*}{4.0}   &  \multirow{2}{*}{3.0}	\\ & Candidate & \makecell[l]{A video artist who uses YouTube and a storyteller who uses storytelling techniques are among \\the nominees for the 2014 Turner Prize.} & & & & & & & & \\ \hline
    \multirow{2}{*}{44} & Reference & \makecell[l]{DJ Dave Lee Travis has told a court he does not have a "predatory nature".}	&  \multirow{2}{*}{ref2cand}   &  \multirow{2}{*}{4.0}   &  \multirow{2}{*}{3.5}   &  \multirow{2}{*}{4.0}   &  \multirow{2}{*}{4.0}   &  \multirow{2}{*}{4.0}   &  \multirow{2}{*}{4.0}   &  \multirow{2}{*}{4.0}	\\ & Candidate & \makecell[l]{Former radio DJ Dave Lee Travis has told a court he is "cuddly" not "predatory".} & & & & & & & & \\ \hline
    \multirow{2}{*}{45} & Reference & \makecell[l]{Naturalist Sir David Attenborough and the Queen are the greatest living British man and woman, \\according to readers of Best of British magazine.}	&  \multirow{2}{*}{cand2ref}   &  \multirow{2}{*}{3.0}   &  \multirow{2}{*}{3.0}   &  \multirow{2}{*}{3.0}   &  \multirow{2}{*}{3.0}   &  \multirow{2}{*}{4.0}   &  \multirow{2}{*}{4.0}   &  \multirow{2}{*}{3.0}	\\ & Candidate & \makecell[l]{Sir David Attenborough has been named the best living British celebrity in a poll by the \\Magazine of British History.} & & & & & & & & \\ \hline
    \multirow{2}{*}{46} & Reference & \makecell[l]{A Chinese woman has been found guilty of trespassing at President Donald Trump's \\Mar-a-Lago club in Florida and of lying to a federal agent.}	&  \multirow{2}{*}{ref2cand}   &  \multirow{2}{*}{2.0}   &  \multirow{2}{*}{5.0}   &  \multirow{2}{*}{3.0}   &  \multirow{2}{*}{4.5}   &  \multirow{2}{*}{1.0}   &  \multirow{2}{*}{1.0}   &  \multirow{2}{*}{1.0}	\\ & Candidate & \makecell[l]{A woman who sparked alarm at Mar-a-Lago has been found guilty of killing herself.} & & & & & & & & \\ \hline
    \multirow{2}{*}{47} & Reference & \makecell[l]{Graduates from ethnic minorities in Britain are less likely to be in work than their white peers, \\a study says.}	&  \multirow{2}{*}{ref2cand}   &  \multirow{2}{*}{5.0}   &  \multirow{2}{*}{3.0}   &  \multirow{2}{*}{3.0}   &  \multirow{2}{*}{3.0}   &  \multirow{2}{*}{4.0}   &  \multirow{2}{*}{5.0}   &  \multirow{2}{*}{5.0}	\\ & Candidate & \makecell[l]{Black and ethnic minority graduates are less likely to be employed than white British counterparts, \\a report suggests.} & & & & & & & & \\ \hline
    \multirow{2}{*}{48} & Reference & \makecell[l]{An anonymous letter sent to a council outlining an alleged plan to oust head teachers is "defamatory", \\the leader of Birmingham City Council has said.}	&  \multirow{2}{*}{ref2cand}   &  \multirow{2}{*}{2.0}   &  \multirow{2}{*}{4.5}   &  \multirow{2}{*}{4.0}   &  \multirow{2}{*}{3.5}   &  \multirow{2}{*}{3.0}   &  \multirow{2}{*}{2.0}   &  \multirow{2}{*}{3.0}	\\ & Candidate & \makecell[l]{A letter written by a council officer calling for schools to be taken over by a council has been defamatory.} & & & & & & & & \\ \hline
    \multirow{2}{*}{49} & Reference & \makecell[l]{The 2017 Oscar nominations are out, with La La Land the frontrunner . Here's a round-up of the \\surprises and talking points from this year's list.}	&  \multirow{2}{*}{ref2cand}   &  \multirow{2}{*}{3.0}   &  \multirow{2}{*}{4.0}   &  \multirow{2}{*}{3.0}   &  \multirow{2}{*}{4.0}   &  \multirow{2}{*}{3.0}   &  \multirow{2}{*}{4.0}   &  \multirow{2}{*}{4.0}	\\ & Candidate & \makecell[l]{The full list of Oscar nominations has been announced. Here are 10 talking points from the shortlists.} & & & & & & & & \\ \hline
    \multirow{2}{*}{50} & Reference & \makecell[l]{People on Jersey's Ecrehous islands are concerned travellers are arriving from France by boat and \\not being tested for coronavirus.}	&  \multirow{2}{*}{ref2cand}   &  \multirow{2}{*}{3.0}   &  \multirow{2}{*}{5.0}   &  \multirow{2}{*}{4.0}   &  \multirow{2}{*}{5.0}   &  \multirow{2}{*}{4.0}   &  \multirow{2}{*}{3.0}   &  \multirow{2}{*}{4.0}	\\ & Candidate & \makecell[l]{People living on Jersey's Ecrehous islands have said they fear they are "playing Russian roulette" \\with coronavirus restrictions after a rise in arrivals.} & & & & & & & & \\
    \bottomrule
  \end{tabular}}
  \tiny
\end{threeparttable}
  \caption{Qualitative analysis of correctness with 50 random samples (Part 2).}
  \label{tab:random-check-sample-p2}
\end{table}

\clearpage
\subsection{Interaction with GPT models in Reference-based Task}
\label{sec:appendix-gpt}
\subsubsection{Prompt Design}
\begin{figure}[!h]
  \centering
  \includegraphics[width=0.8\textwidth]{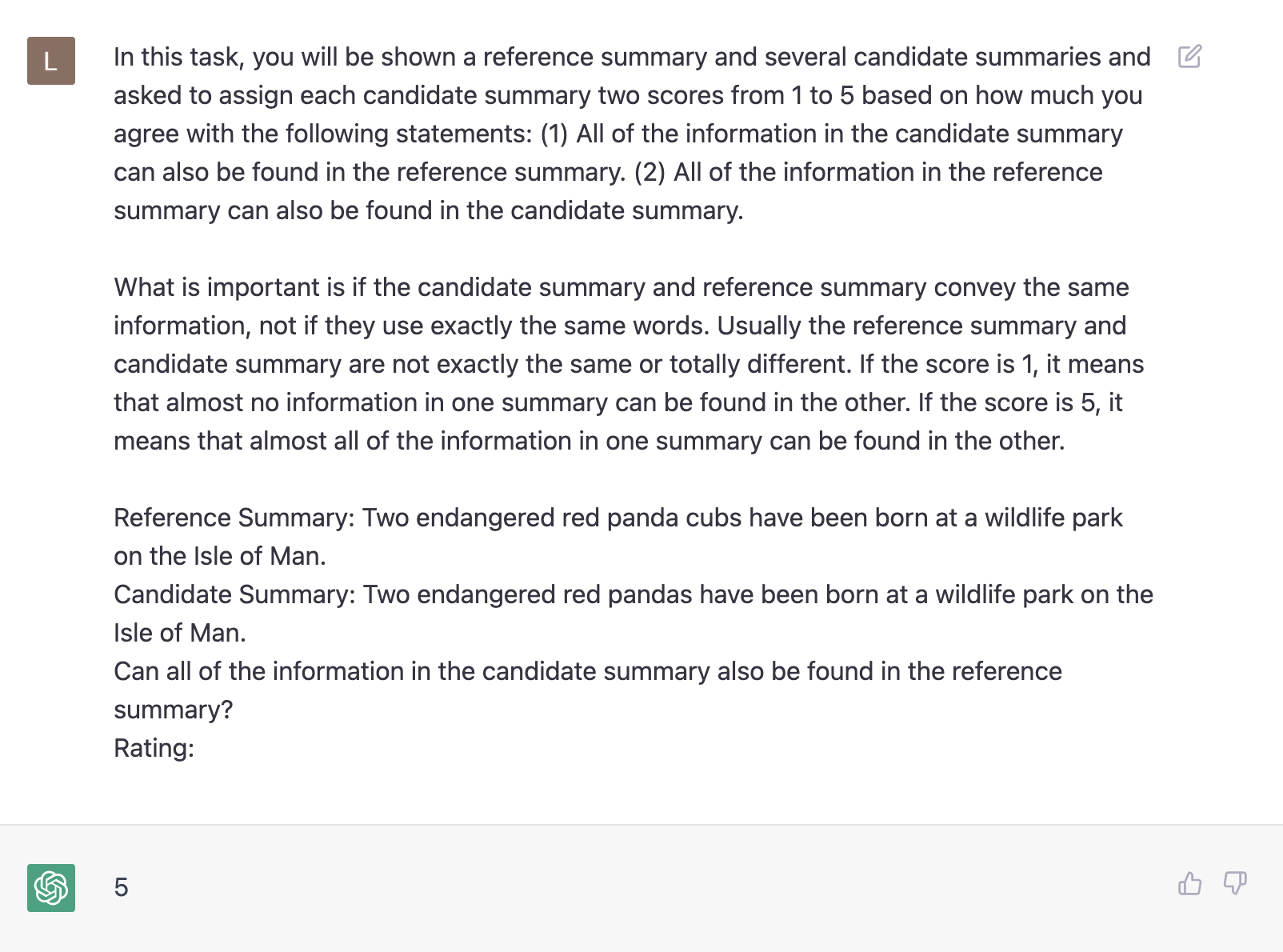}
  \caption{Example of interaction with ChatGPT in the reference-based task.}
  \label{fig:chatgpt-instruct}
\end{figure}

In Figure \ref{fig:chatgpt-instruct}, we show an example of the interaction with ChatGPT and the exact prompt design we use to acquire scores generated by GPT models through API\footnote{\url{https://platform.openai.com/docs/api-reference}} for the analysis of correctness in the reference-based task.

This prompt design follows the instructions we provide to the crowd annotators in the reference-based task (see Figure \ref{fig:ref-instruction} for details) with minor modifications for the score generation from GPT models. Details about running experiments through API can be found in Section \ref{subsec:refbased-results}.

\subsubsection{Estimated Cost of GPT Models}
\label{sec:app-gpt-cost}
We estimate the cost of running GPT models for the score generation in the reference-based task (240 annotation questions in total) based on the cost of 50 random annotation questions. Details of pricing can be found on OpenAI's website\footnote{\url{https://openai.com/pricing}}. We assume the GPT model only returns the score without explanations. 

\begin{table}[!h]
\small
\centering
\begin{threeparttable}
  \centering
  \resizebox{0.45\textwidth}{!}{
  \begin{tabular}{ccc}
    \toprule
    GPT Models
    & Cost per 1K Token
    & Estimated Cost \\
    \hline
    GPT-3.5 & \$0.02 & \$0.21 \\
    ChatGPT & \$0.002 & \$0.02 \\
    GPT-4 & \makecell{\$0.03 (prompt)\\\$0.06 (completion)} & \$0.32 \\
    \bottomrule
  \end{tabular}}
  \caption{Estimated cost of GPT models for the reference-based task.}
  \label{tab:gpt-cost}
  \end{threeparttable}
\end{table}

\clearpage
\subsection{Instruction and Annotation Question Examples of HIT}
\label{sec:app-instructions}
Here we provide some examples of instructions and annotation questions for all three tasks as screenshots. 
% More details can be found at this link.\footnote{We will add the link to the website with sample HITs here later.}
% \footnote{\url{https://www.seas.upenn.edu/~ddeutsch/gem/2022-06-27/gem-v2-summarization-qualification/index.html}}.

\subsubsection{Qualification Task}
\begin{figure}[!htb]
  \centering
  \includegraphics[width=\textwidth]{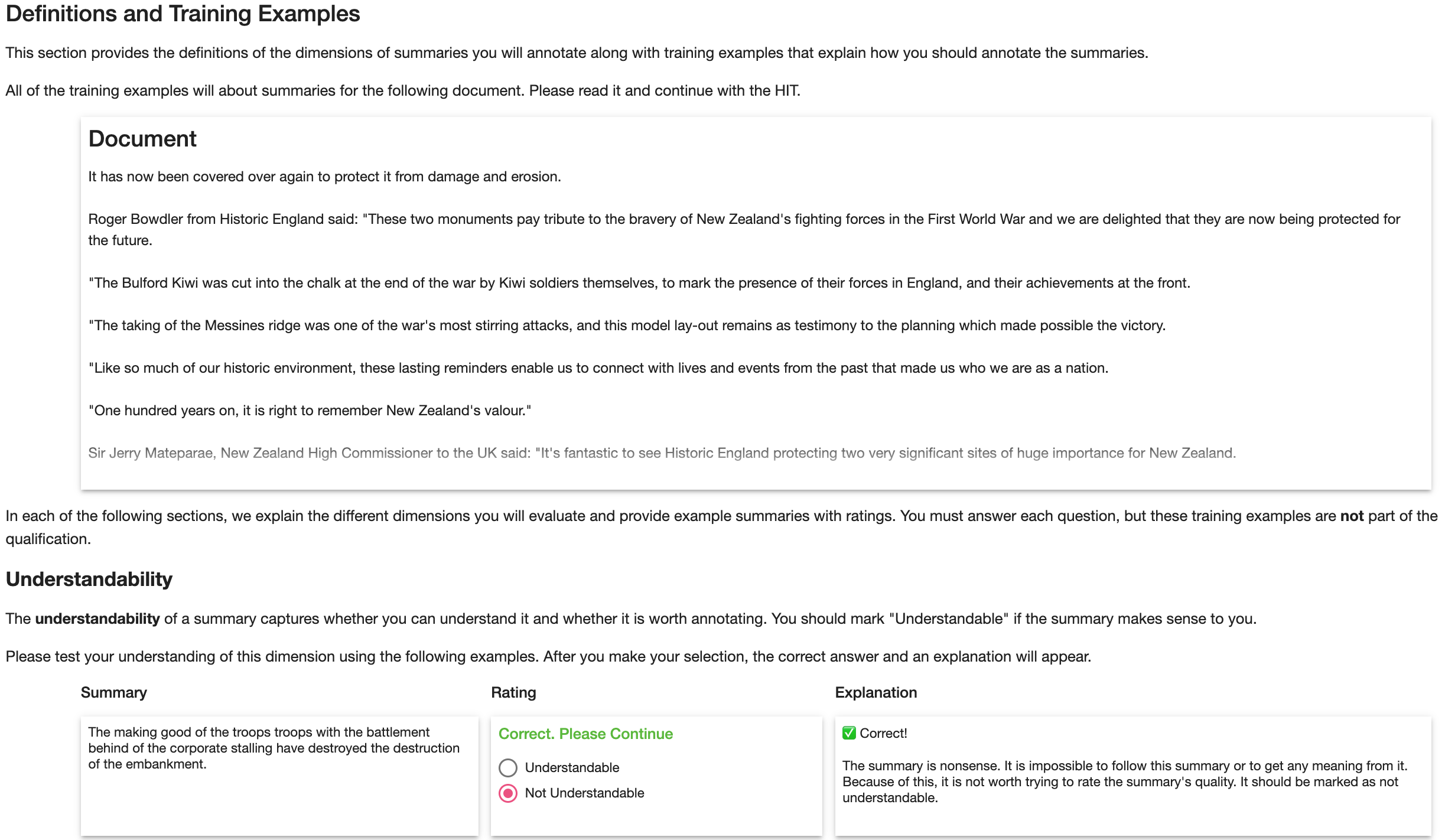}
  \caption{Example from training part of qualification task.}
  \label{fig:qual-train-example}
\end{figure}

\begin{figure}[!htb]
  \centering
  \includegraphics[width=\textwidth]{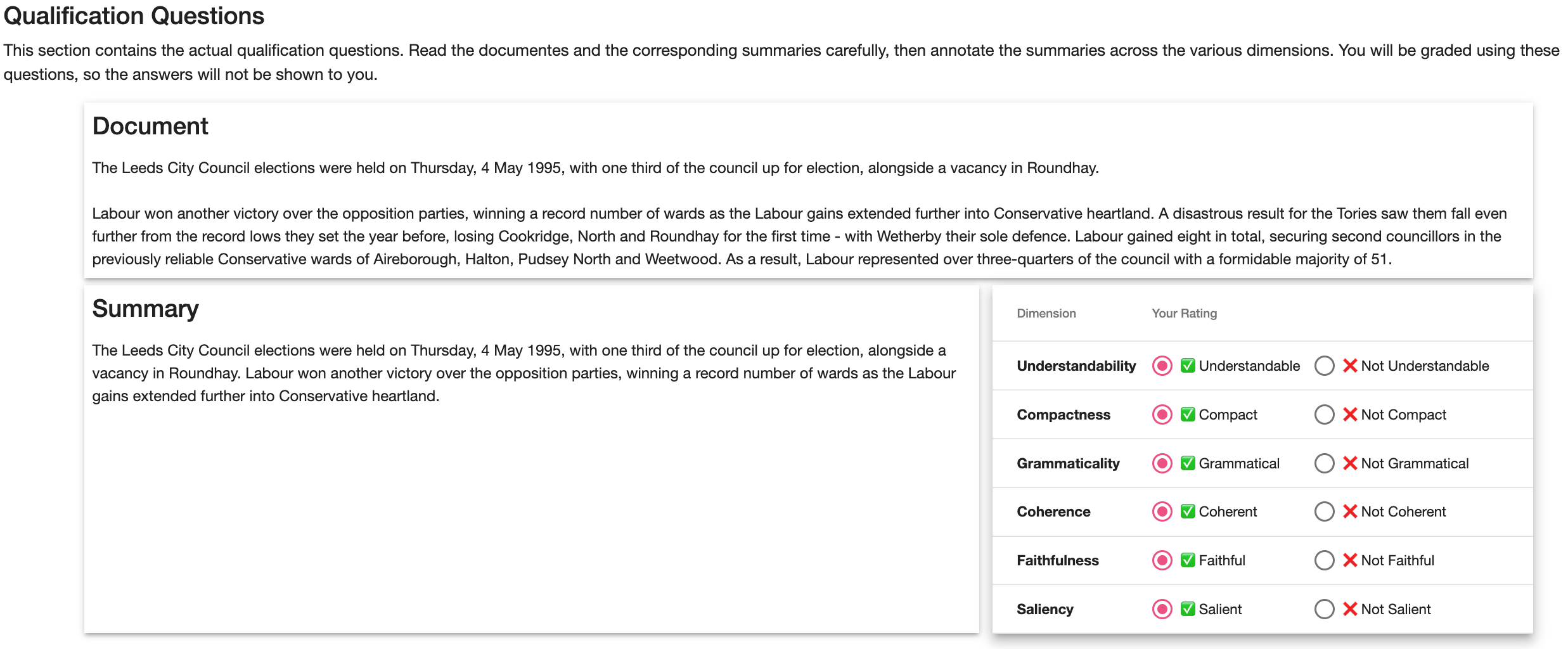}
  \caption{Example from qualification part of qualification task.}
  \label{fig:qual-question}
\end{figure}

\begin{itemize}
    \item Figure \ref{fig:qual-train-example} shows the definition of an evaluation dimension illustrated with examples in the training part. 
    \vspace*{-3mm}
    \item Figure \ref{fig:qual-question} shows the example of the qualification question in the qualification part. 
\end{itemize}

\clearpage
\subsubsection{Endurance Task}
\begin{figure}[!htb]
  \centering
  \includegraphics[width=\textwidth]{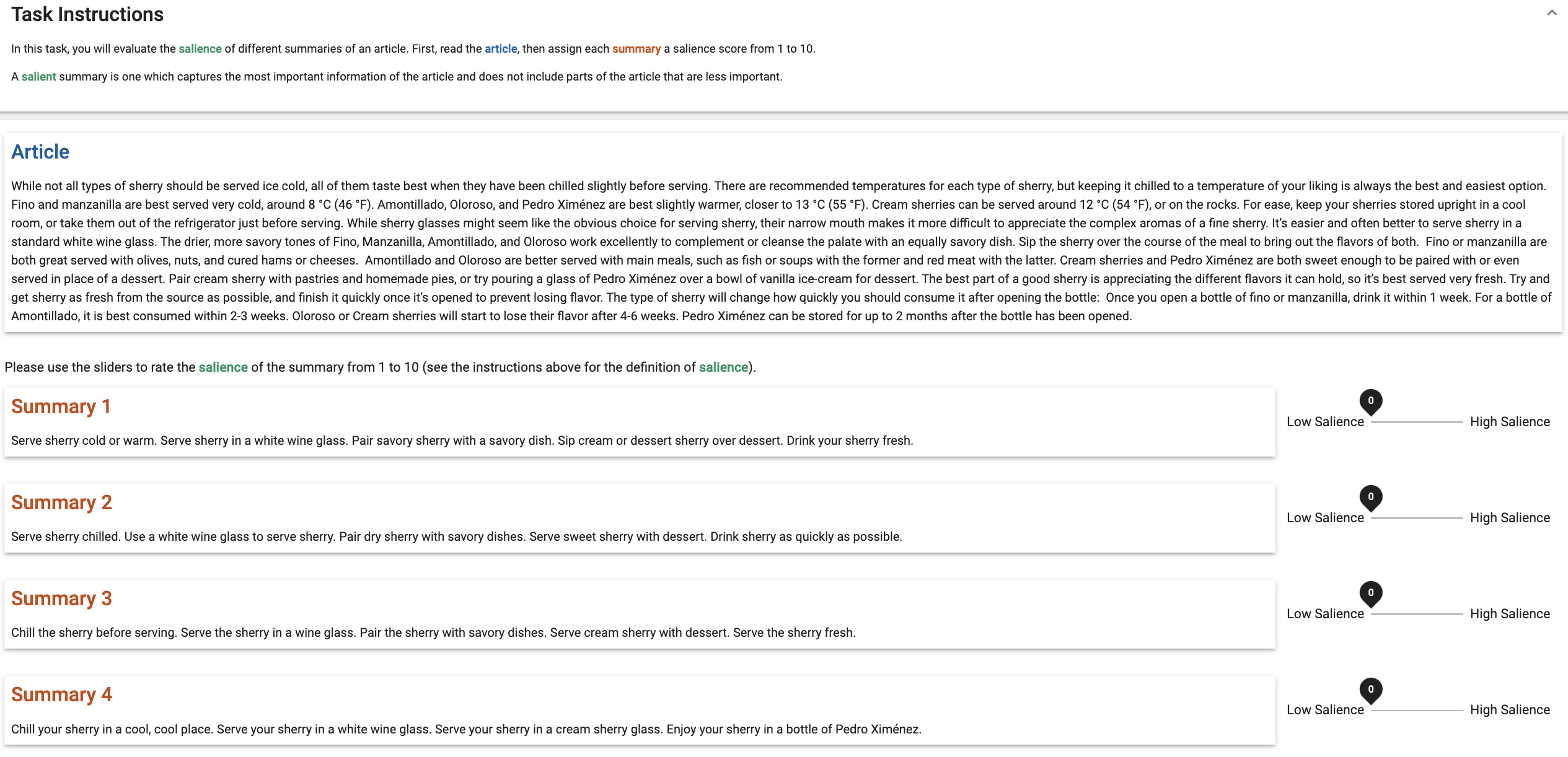}
  \caption{Example of the annotation question in endurance task.}
  \label{fig:endu-question}
\end{figure}

Figure \ref{fig:endu-question} shows the example of the annotation question on a Likert scale of 1 to 10 in the endurance task.

\subsubsection{Reference-based Task}
\begin{figure}[!htb]
  \centering
  \includegraphics[width=\textwidth]{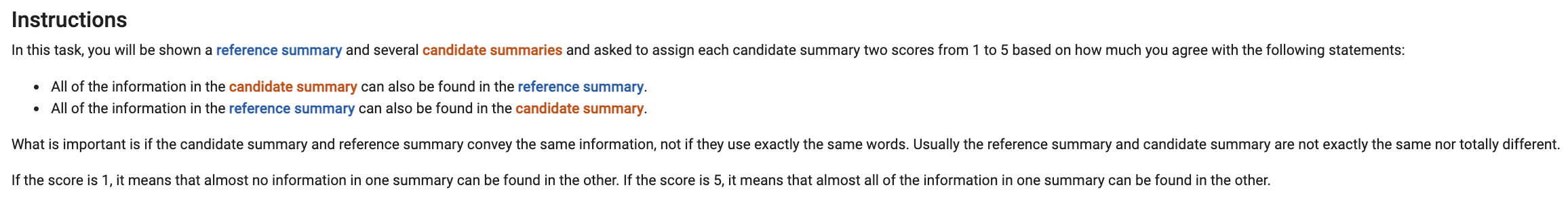}
  \caption{Instructions for the reference-based task.}
  \label{fig:ref-instruction}
\end{figure}

\begin{figure}[!htb]
  \centering
  \includegraphics[width=\textwidth]{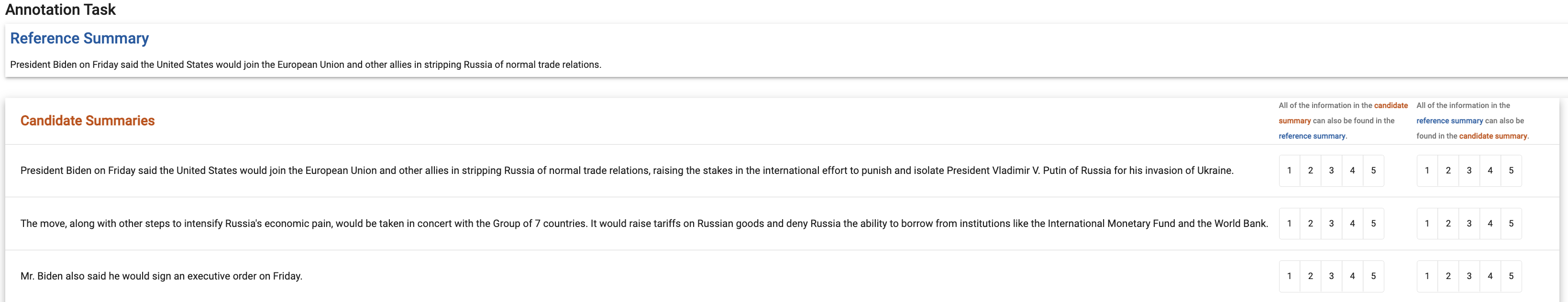}
  \caption{Example of the annotation question in the reference-based task.}
  \label{fig:ref-question}
\end{figure}

\begin{itemize}
    \item Figure \ref{fig:ref-instruction} shows the instructions for the reference-based task.
    \vspace*{-3mm}
    \item Figure \ref{fig:ref-question} shows the example of the annotation question in the reference-based task.
\end{itemize}

% \clearpage
% \input{Human-Eval-Datasheet/human-evaluation-datasheet.tex}
% \import{./}{human-evaluation-datasheet.tex}
% \thispagestyle{acl2023}

\end{document}